\PassOptionsToPackage{table}{xcolor}

\documentclass[10pt,twocolumn,letterpaper]{article}

\usepackage[pagenumbers]{cvpr} 

\usepackage[frozencache,cachedir=minted-cache]{minted}

\usepackage[accsupp]{axessibility}  

\usepackage{xfrac}
\usepackage{xspace}
\usepackage{multirow}
\usepackage{algorithm}
\usepackage{algpseudocode}
\usepackage{mathtools}

\usepackage{bm}
\usepackage{xcolor}

\usepackage{pgfplots}

\everymath=\expandafter{\the\everymath\displaystyle}

\newcommand{\pname}[1]{\textit{#1}}
\newcommand{\dname}[1]{\textit{#1}}
\newcommand{\raft}[0]{\pname{RAFT}\xspace}

\DeclarePairedDelimiter\ceil{\lceil}{\rceil}
\DeclarePairedDelimiter\floor{\lfloor}{\rfloor}

\algblockdefx[CUDAKERNEL]{CudaKernel}{EndCudaKernel}[1]{{\textbf{On Device} #1}}{{\textbf{End Device}}}

\definecolor{cvprblue}{rgb}{0.21,0.49,0.74}
\usepackage[pagebackref,breaklinks,colorlinks,allcolors=cvprblue]{hyperref}

\usepackage{placeins}
\usepackage{paralist} 

\usepackage{listings}

\title{
Efficient All-Pairs Correlation Volume Sampling for Optical Flow Estimation
}

\author{
  Karlis Martins Briedis$^{1,2}$ \qquad
  Markus Gross$^{1,2}$ \qquad
  Christopher Schroers$^{1}$
  \vspace{0.5em} \\
  {$^1$DisneyResearch\textbar Studios} \qquad
  {$^2$ETH Z\"urich}
\vspace{4mm}
}

\begin{document}
\maketitle

\begin{abstract}
Recent optical flow estimation methods often employ local cost sampling from a dense all-pairs correlation volume. This results in quadratic computational and memory complexity in the number of pixels. Although an alternative memory-efficient implementation with \textit{on-demand} cost computation exists, this is significantly slower in practice and
therefore many prior methods process images at downsampled resolutions, missing fine-grained details.

To address this, we propose an algorithm for both memory and compute-efficient
implementation of the all-pairs correlation volume sampling,
still matching the exact mathematical operator as defined by RAFT.
Our approach outperforms \textit{on-demand} sampling by up to $92\%$ while maintaining equally low memory usage,
and performs at least on par with the default implementation with up to $99\%$ lower memory usage.
As cost sampling makes up a significant portion of the overall runtime, this can translate to up to $63\%$ savings for the total end-to-end model inference on high-resolution inputs.
Our evaluation of existing methods includes an \textit{8K} ultra-high-resolution dataset and an
inference-time extension of the SEA-RAFT method. With this, 
we achieve state-of-the-art results at high resolutions both in accuracy and runtime.
\end{abstract}

\section{Introduction}
\label{sec:intro}

\begin{figure}[t]
  \centering
  \noindent
  \begin{minipage}{\linewidth}
    \centering
    \resizebox{\linewidth}{!}{\input{include/kvqtptoxlrdsrgofcczg.pgf}
    }
  \end{minipage}

\vspace*{-2mm}
  \caption{
  Comparison of the top-performing optical methods on an 8K ultra-high-resolution dataset.
  We plot the lowest 1px error and the respective runtime for each
  method across all inference resolutions.
  Additionally, we show the runtime improvement achieved by
  using our correlation sampling algorithm in \pname{RAFT}-based methods
  with green arrows.
}
  \label{fig:teaser}
   \vspace*{-3mm}
\end{figure}

Optical flow estimation is a classical low-level computer vision problem
that involves estimating dense correspondences between video frames.
It has found applications in many downstream video tasks,
including
action recognition~\cite{sunOpticalFlowGuided2018},
video compression~\cite{agustssonScaleSpaceFlowEndtoEnd2020},
video inpainting~\cite{kimDeepVideoInpainting2019, xuDeepFlowGuidedVideo2019},
and frame interpolation~\cite{niklausContextAwareSynthesisVideo2018}.
Many of these tasks are used to process ultra-high-resolution (UHR)
content, such as in movie post-production, requiring high-quality flows
at their original resolution.

The vast majority of optical flow estimation methods use
cost matching between two input images~\cite{xuAccurateOpticalFlow2017,sun2018pwc},
with recent methods adopting dense all-pairs correlation volume sampling
based on \raft~\cite{teedRAFTRecurrentAllPairs2020}.
For correlation volume sampling,
\raft either pre-computes the full 4D volume or
uses a memory-efficient \textit{on-demand} cost sampling with
a custom \pname{CUDA} implementation.
Both options have been adopted by many subsequent
works~\cite{jahediCCMRHighResolution2024,jahediMSRAFTHighResolution2024,jiangLearningEstimateHidden2021,morimitsuDPFlowAdaptiveOptical2025}.
However, existing implementations encounter issues
with high-resolution inputs.
The full volume computation complexity grows quadratically with respect
to the number of pixels, making it prohibitive to store at high resolutions.
In contrast, the \textit{on-demand} sampling achieves memory reduction at the
expense of inefficient computations thus resulting in worse runtime performance.
Due to this trade-off, several recent methods have been proposed to avoid
sampling from the full volume~\cite{xuHighResolutionOpticalFlow2021,zhaoHybridCostVolume2024},
or removing it~\cite{Kiefhaber_2025_ICCV,wangWAFTWarpingAloneField2025}, but
typically do so at the expense of estimation accuracy.

In this work, we propose a novel all-pairs correlation volume sampling algorithm,
based on computation and sampling of a partial block sparse cost volume.
By combining the strengths of both previous approaches, it eliminates the need
for approximations
or trade-off between speed and memory, and can be cheaply used even at very
high input resolutions.
The algorithm design is motivated by our analysis of the typical sampling patterns
of the full correlation volume in practical applications,
where we observe that only a small, regular part of the volume is sampled.
This enables us to only compute the necessary subsections of the volume efficiently.

In isolation, the operation performs
at least on par with the default implementation,
while having up to $99\%$ lower memory usage,
and it outperforms \textit{on-demand} sampling by up to $92\%$ while
at equally low memory usage.
When used in \raft as a drop-in replacement, it reduces the total end-to-end
runtime by up to $67\%$ compared to existing methods that are feasible to
run with modern hardware.
When applied to \pname{SEA-RAFT}~\cite{wangSEARAFTSimpleEfficient2025},
which is already designed for efficiency,
it provides approximately $33\%$ runtime reduction.

Additionally, we generate a realistic
\textit{8K} optical flow
dataset based on the \textsc{Blender} movie \textsc{Charge}
and use it to evaluate existing optical flow methods
at ultra-high-resolutions.
Observing that estimating optical flow at high resolutions
improves fine-grained details at the expense of not 
capturing large displacements,
we propose a test-time extension of the
\pname{SEA-RAFT} method to perform cascaded inference.
Without any additional training of the model, it allows
reducing the endpoint error for large motion.
Combined with our efficient correlation sampler, we achieve
state-of-the-art results at the Pareto front of accuracy and
runtime for \textit{8K} flow estimation (Figure~\ref{fig:teaser}).

\begin{figure*}[t]
  \centering
  \includegraphics[width=0.99\linewidth]{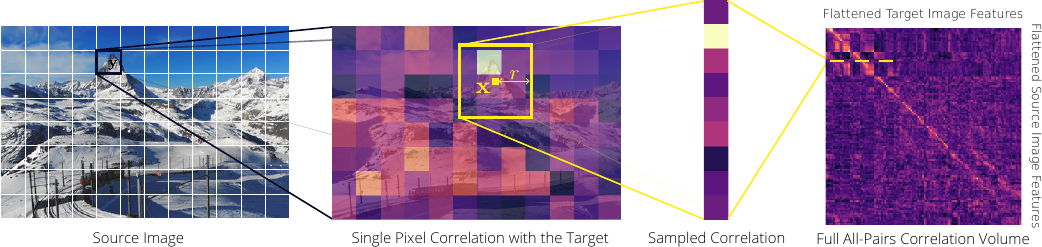}
  \vspace*{-1.5mm}
  \caption{
  Overview of the correlation volume sampling.
  Given a map of correlation between the features of a single source pixel and
  the features of
  every pixel in another image, bilinear sampling is used to extract local matching
  costs around a point of interest.
  When repeated for every source pixel, the costs are stored in a dense all-pairs
  correlation volume, where each row and column correspond to a source and target pixel, respectively.
  This is repeated on multiple levels of scale.
  }
  \label{fig:problem}
  \vspace*{-1.5mm}
\end{figure*}
 
\section{Related Work}
\label{sec:related}
\paragraph{Optical Flow Estimation.}
Traditional optical flow methods have used variants of local and global
cost volumes.
Here we only cover the methods most closely related to our work
and refer to the survey by Zhai~\textit{et al.}~\cite{zhaiOpticalFlowScene2021}
for a more complete overview.

\pname{FlowNetC}~\cite{dosovitskiyFlowNet2015} and \pname{PWC-Net}~\cite{sun2018pwc} reintroduce the classical concept of cost volume computation for deep learning applications, but process a flattened local cost between source and warped target images.
Similarly, several other methods employ cost volume processing but limit processing
to the local neighborhood~\cite{hurIterativeResidualRefinement2019, yangVolumetricCorrespondenceNetworks2019}.

\raft~\cite{teedRAFTRecurrentAllPairs2020} revisits the construction
of an all-pairs correlation volume and combines it with recurrent iterative
refinements that sample matching costs.
It proposes two implementations of 4D correlation volume sampling
that remain commonly used to date.
This approach has subsequently been adopted by many methods,
improving estimation of occluded regions~\cite{jiangLearningEstimateHidden2021},
encoding matching costs~\cite{huangFlowFormerTransformerArchitecture2022, shiFlowFormerMaskedCost2023},
diffusion-based flow updates\cite{luoFlowDiffuserAdvancingOptical2024},
and other improvements~\cite{luoGAFlowIncorporatingGaussian2023, luoLearningOpticalFlow2022,suiCRAFTCrossAttentionalFlow2022, sunSKFlowLearningOptical2022, zhaoGlobalMatchingOverlapping2022,stoneSMURFSelfTeachingMultiFrame2021}.
\pname{SEA-RAFT}~\cite{wangSEARAFTSimpleEfficient2025} revisits the original
\pname{RAFT} and
introduces simple extensions to improve the efficiency and quality.

Other methods~\cite{jahediCCMRHighResolution2024, jahediMSRAFTHighResolution2024}
distribute the flow updates
over multiple levels up to $\sfrac{1}{2}$ resolution.
Due to sampling matching costs at high resolutions,
the slower \textit{on-demand} sampling has to be used.

Little work has been done on optical flow estimation of high input resolutions.
Xu~\etal~\cite{xuHighResolutionOpticalFlow2021} decompose optical flow
estimation into two 1-dimensional operations, enabling processing up to 4K
images and provide qualitative results of 4K flow estimation.
Similarly, DIP~\cite{zhengDIPDeepInverse2022} targets efficient inference at
high resolution.
The Spring benchmark~\cite{Mehl2023_Spring}
increases the realism and resolution of optical flow evaluation
but is limited to HD resolution ($1920\times1080$px).
GMFlow~\cite{xuGMFlowLearningOptical2022} considers global motion
but consequently becomes prohibitively expensive at high resolutions.
Morimitsu~\etal~\cite{morimitsuDPFlowAdaptiveOptical2025} target high resolution
inference via an adaptive flow architecture and concurrently introduce
quantitative evaluations on a synthetic dataset up to \textit{8K} resolution.

Several works~\cite{garrepalliDIFTDynamicIterative2023a,jiangLearningOpticalFlow2021,xuHighResolutionOpticalFlow2021,zhaoHybridCostVolume2024}
focus on reducing computational costs and memory usage by proposing
architecture changes that avoid sampling the dense correlation volume
but do so at the expense of estimation accuracy.
More recently, ReCoVEr~\cite{Kiefhaber_2025_ICCV} and
WAFT~\cite{wangWAFTWarpingAloneField2025}
remove cost volumes to avoid their time and memory complexity,
and compensate it with use of larger feature extraction backbones.

Unlike prior work, we propose an algorithm to improve
the efficiency of all-pairs correlation sampling, which is directly
compatible with \pname{RAFT} and its variants with no compromise in quality,
and show that cost volumes can be used without their commonly assumed
computational and space costs.

\paragraph{Operator Efficiency Improvements.} Several 
prior works have focused on improving the computational efficiency
of commonly used deep learning operators.
FlashAttention~\cite{dao2022flashattention, daoFlashAttention2FasterAttention2023} 
proposes a memory-efficient implementation of attention as introduced in the 
Transformer~\cite{vaswaniAttentionAllYou2017}.
Neighborhood~Attention~Transformer~\cite{hassaniNeighborhoodAttentionTransformer2023}
provides an efficient implementation for local attention computation and proposes a
method utilizing the efficient operator.
In contrast, we not only propose a more efficient implementation of the all-pairs
correlation volume sampling, matching the exact mathematical operator as defined
by {\raft},
but also introduce high-level algorithm improvements to the correlation sampling process.
 
\section{Correlation Volume Sampling Analysis}
\label{sec:analysis}

In this section, we introduce the correlation sampling problem
in more detail,
present the default implementation, and
provide an analysis of access patterns of the 
full 4D correlation volume, which form the foundation of our algorithm.
The terms \textit{cost} and \textit{correlation} volume are used interchangeably.

\subsection{Problem and Baseline Implementation}

Given $D$-dimensional features $F^{1, 2} \in \mathbb{R}^{H \times W \times D}$ extracted from two images,
the visual similarity, or correlation, between two pixels
is defined as the dot product of their feature vectors.

A lookup is performed by bilinearly sampling at a local grid around an interest pixel
$\mathbf{x}$,
with sampling positions defined as integer offsets within radius $r$.
More formally, the sampled correlation at source pixel $\mathbf{y}$ is defined as
\begin{equation}
\label{eq:correlation}
\mathcal{C}_r(\mathbf{y}, \mathbf{x}) = \{\langle F^1_{\mathbf{y}}, F^2_{\mathbf{x} +\mathbf{dx}} \rangle \vert \mathbf{dx} \in \mathbb{Z} \land ||\mathbf{dx}||_\infty \leq r \}.
\end{equation}
See \cref{fig:problem} for a visualization.

The default implementation of correlation sampling first precomputes a dense
4-dimensional correlation 
volume $\mathbf{C} \in \mathbb{R}^{H_1 \times W_1 \times H_2 \times W_2}$, where $H$ and $W$ are the height and width of both image 
features.
This is achieved by flattening both images along spatial dimensions,
as illustrated on the right side of \cref{fig:problem}, and
computing the full correlation volume using a single matrix-matrix multiplication 
\begin{equation}
\label{eq:dense_volume}
\begin{gathered}
\mathbf{C} = \bar{F}^1 \cdot \bar{F}^2,\\ \text{where} \ \bar{F}^1 \in \mathbb{R}^{[H_1 \times W_1] \times D}, \bar{F}^2 \in \mathbb{R}^{[H_2 \times W_2] \times D}.
\end{gathered}
\end{equation}
The output is then reshaped back to 4 dimensions, and
bilinear sampling is directly performed on the precomputed~$\mathbf{C}$.

In practice, \raft constructs a 4-level pyramid by average pooling the last two 
dimensions of $\mathbf{C}$ and performs a lookup on the pooled volumes to increase
the perceptual window.

Alternatively, a memory-efficient \textit{on-demand} implementation computes the 
values of Eq.~\ref{eq:correlation} directly for each source frame pixel.
This approach reduces computational and memory complexity and does not require storing 
any intermediate values.
However, in practice, it underperforms compared to the baseline
due to operations that are not hardware-friendly and the lack of result reuse
between iterations. 
We refer to \raft~\cite{teedRAFTRecurrentAllPairs2020} 
for more details on the sampling procedure.

\subsection{Correlation Volume Access Patterns}
\label{sec:analysis:access_patterns}

The lookup grid for each update
is defined over a local neighborhood around the
current flow estimate.
It ensures that the number of sampled cells
per row is limited to 
the local grid size, \textit{i.e.}, $(2r+1)^2$ elements.
During iterative updates,
the local neighborhood is shifted but remains close to the 
previous iteration with a significant overlap of the
sampled region.
Thus, across all flow update iterations, the total number of sampled 
columns per single source pixel remains low and does not increase
with input size.

\begin{figure}[tb]
  \centering
  \includegraphics[width=0.9\linewidth]{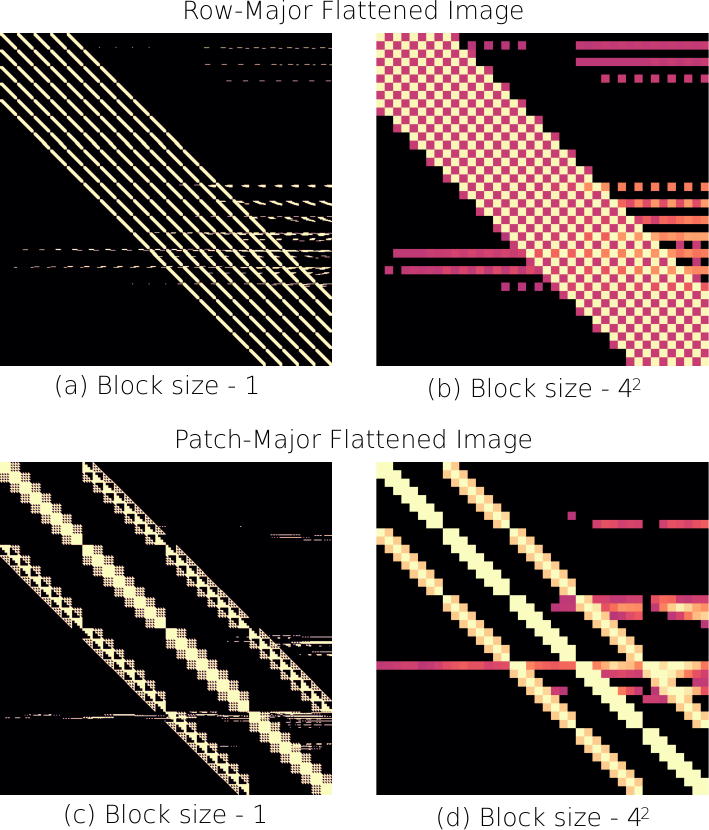}
    \vspace{-1mm}
  \caption{
  Sampling patterns of a single image over all \raft iterations.
  Dark regions correspond to cells that have not been sampled while
  lighter values indicate more sampled values per block.
  }
  \label{fig:sampling_analysis}
  \vspace{-2mm}
\end{figure}

To empirically verify this, we run the default \raft implementation
on the \pname{Sintel}~\cite{Butler:ECCV:2012} optical flow training dataset,
while tracking which correlation volume cells are sampled.
In Fig. \ref{fig:sampling_analysis}[a], we visualize the sampled cells for a single image over all
update iterations, represented as a 
$[H_1 \times W_1] \times [H_2 \times W_2]$ volume,
where each row shows all matches of a single source pixel.
On average, over the whole dataset, only $1.6\%$ of the cells are being sampled.
This suggests that we can build a more efficient algorithm that only
computes the necessary elements.

However, keeping track of which values are required and
computing them efficiently is challenging.
A straightforward way is to track them in a binary mask, but
that would consume as much memory as the full correlation volume,
thus becoming infeasible, while the computations would be similar to that of
\textit{on-demand} sampling.

To improve that, we suggest representing $\mathbf{C}$ in a block sparse format
and making decisions on computations per block rather than per pixel.
As shown in Fig. \ref{fig:sampling_analysis}[b],
averaging the number of sampled values per block
slightly increases the ratio of cells 
that need to be computed but remains significantly lower than
the full matrix for sufficiently small block sizes.
Full numerical results of the sampling patterns are provided in the supplementary material.

\paragraph{Reshaped Cost Volume.} As the local sampling
grid is defined over a 2D neighborhood,
when flattened, the values for each row are scattered over multiple
column groups and span over many blocks when averaged
(Fig. \ref{fig:sampling_analysis}[a,b]).
This can be improved by pooling the cost volume in a different layout
that groups cells based on 2D patches on both source and target images.

To this end, we reshape the input images into a \textit{patch-major} format,
where each block is represented continuously in memory, before
computing the correlation volume. As shown in Fig. \ref{fig:sampling_analysis}[c,d],
the block-aware layout significantly increases sparsity
without any computational overhead.

\label{sec:access_patterns}

\section{Efficient Correlation Volume Sampler}
\label{sec:method}

Based on our observations on sampling patterns made in Section~\ref{sec:analysis}, we propose an efficient algorithm for all-pairs correlation sampling.

\begin{figure*}[tb]
  \centering
  \includegraphics[width=0.99\linewidth]{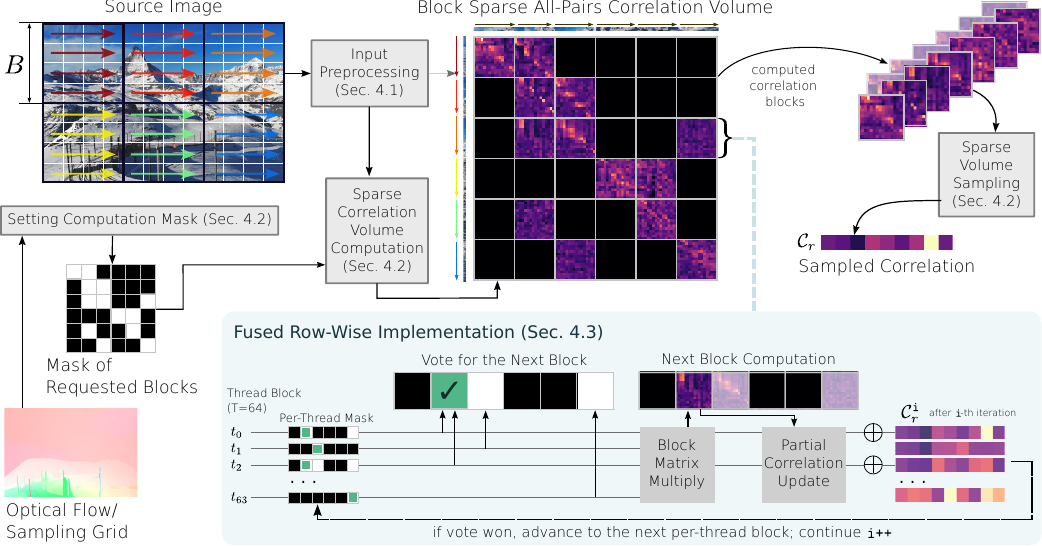}
  \caption{
  Algorithm overview. It consists of input preprocessing and 3 steps per iteration:
\textit{a}) determining blocks that need to be computed;
\textit{b}) computing selected blocks with block sparse matrix-matrix multiplication;
\textit{c}) sampling computed blocks.
  }
  \label{fig:method}
  \vspace*{-3mm}
\end{figure*}

An overview of the algorithm is shown in~Fig~\ref{fig:method}.
Inputs are pre-processed once per given a pair of images
and on every iteration of \pname{RAFT}-based flow updates
(typically $4-32$ iterations)
we perform three main steps:
\begin{inparaenum}[\itshape a\upshape)]
\item determining regions of the volume that will get sampled and
setting the computation mask;
\item computing selected blocks with efficient tiled matrix-matrix multiplications;
\item sampling computed blocks.
\end{inparaenum}

First, we describe a high-level algorithm, where computation mask and block
sparse all-pairs correlation volume is stored explicitly.
Then, we present a fused implementation that avoids building the
computation mask and stores one block at a time, achieving linear
$\mathcal{O}(n)$ time and space complexity in the number of input pixels.

We only cover the single-level case, which is extended to
multi-level correlation volumes by computing every level independently,
taking average-pooled target features $F^2$.

More implementation details, pseudocode,
and complexity derivation are provided in the supplementary material.

\subsection{Input Preprocessing}

To minimize the number of blocks that need to be computed, we store images in
a \textit{patch-major} format
as described in Sec.~\ref{sec:analysis:access_patterns} with block height $B$.
To simplify the algorithm implementation, we only consider rectangular blocks and 
use blocks of the same size across all steps.

At first, we pad inputs to a multiple of $B$, and split the image into
$B^2$-sized tiles.
Each tile is then independently flattened in \textit{row-major} order followed by 
all tiles being flattened in \textit{row-major} order.
This is visualized in Fig.~\ref{fig:method} with different-colored arrows.
The flattened image is then stored in a contiguous memory block.

\subsection{High-Level Implementation}
\label{sec:method:high_level}

\paragraph{Setting Computation Mask.} First, we build a mask of which cell blocks in the cost volume need to 
be computed, such that all necessary values can be sampled.
To this end, we take the convex integer grid positions for the original problem,
as defined in Eq.~\ref{eq:correlation}, and then floor divide them with the
chosen block size $B$ to obtain the cell indices in the block mask.
We then perform the reverse operation of grid sampling - \textit{i.e.}
scattering - to set these cells as $1$ in the mask,
while the rest is initialized with $0$.
By performing the reverse operation of the sampling step, we ensure that
all sampled locations have been set to be computed.

\paragraph{Sparse Correlation Volume Computation.} The mask is used to compute all blocks of the dense correlation volume
that correspond to the non-zero entries of the mask,
replacing the dense correlation volume computation
defined in Eq.~\ref{eq:dense_volume}.

The computation is performed by a sampled block sparse matrix-matrix
multiplication, where only non-zero blocks are stored in memory.
As each block is produced by computing the product of two $B^4 \times D$
matrices, it is well optimized in hardware and thus can be computed efficiency.

\paragraph{Sparse Volume Sampling.} In order to sample the sparse correlation volume, for each looked up cost value,
the block index is computed as when setting the computation mask.
Then, we gather the block index in the memory, and compute the coordinates
relative to this block.
Finally, the block is locally sampled similar to the baseline.

\subsection{Fused Implementation}

All steps of the algorithm, described in Section~\ref{sec:method:high_level},
can be implemented in a fused \pname{CUDA} kernel, where each row of blocks
of the dense correlation volume is serially computed and directly
sampled in a single \pname{CUDA} thread block, each thread sampling from one row,
\textit{i.e.} computing correlation for one source pixel.
This is visualized in the bottom part of Figure~\ref{fig:method}.
As all rows of blocks are independent, a thread block can perform all steps of
the algorithm without writing any of the intermediate results to the global
memory, but instead storing them in the shared memory and registers.

\paragraph{Implicit Computation Mask.} To avoid storing, setting, and iterating over the full row of computation mask,
we build it implicitly through inter-thread voting of the next non-zero cell.

First, similar to setting an explicit computation mask,
each thread computes indices of all blocks it needs to sample from.
For commonly used parameters, it is no more than $9$ blocks, and we refer
to the complexity analysis in the supplementary for derivation.
As the local grid is well-structured, we can obtain their indices in a strictly
increasing order and store them in a register array.

Next, all threads \textit{vote} for the next smallest block to be computed
within the thread block.
The vote is implemented via an atomic minimum operation to a scalar in the
shared memory, accessible to all threads in a block.
After the vote, all threads fetch the index of the block that needs to be
computed, and advance their local pointers to the next smallest block if 
their vote matched the outcome.

\paragraph{Block Computation and Sampling.} Having agreed on the next tile of the block sparse correlation volume, 
all threads fetch input values and jointly compute the correlation values
(\textit{Block Matrix Multiply} in Fig.~\ref{fig:method}),
similar to how it is done in general matrix multiplications.

After computing the correlation tile,
threads exchange their register values through shared memory,
such that each thread stores all values corresponding to
their source pixel.
These registers are directly sampled to obtain partial sampled correlation
values, adding residual to a global memory buffer
(\textit{Partial Correlation Update} in Fig.~\ref{fig:method}).
After iterating over the whole row of blocks, the output buffer
contains the correct and exact correlation values.
  
\section{Isolated Sampling Evaluation}
\label{sec:results}
\begin{figure*}[t]
    \centering
    \begin{minipage}[b]{0.47\linewidth}
        \centering
        \includegraphics[width=\linewidth]{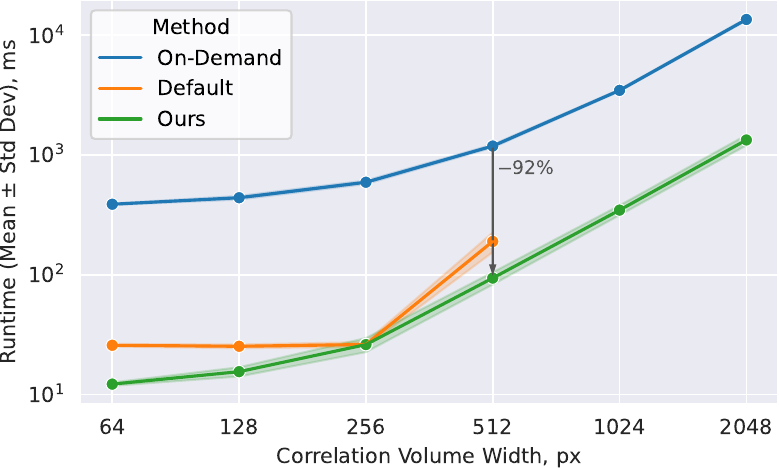}
    \end{minipage}\hfill
    \begin{minipage}[b]{0.47\linewidth}
        \centering
        \includegraphics[width=\linewidth]{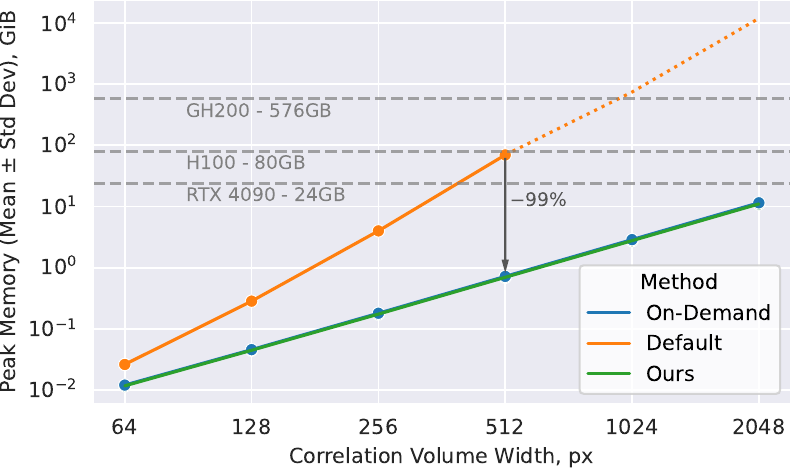}
    \end{minipage}
    \vspace{-2.5mm}
    \caption{Runtime and peak memory consumption depending on the full correlation volume width.
    Standard deviation is displayed as shaded area,
    and we show memory capacity
of different hardware as dotted lines.}
    \label{fig:benchmark_scale}
        \vspace*{-2mm}
\end{figure*}

To evaluate the proposed algorithm, we first conduct experiments in isolation
by only considering the all-pairs correlation sampling.

We run all methods on the \textit{final} pass of
the \pname{Sintel}~\cite{Butler:ECCV:2012} benchmark's train set consisting of $1041$ samples.
The isolated tests are performed by extracting the intermediate query centroids
with \raft at \sfrac{1}{8} resolution and using randomly-generated feature vectors.
For different resolutions, we upsample or downsample the input image.
Unless otherwise noted, the experiments are run using
\textit{PyTorch} \texttt{2.2.2}, \textit{CUDA} \texttt{12.2} and
\textit{NVIDIA GH200} chip, equipped with $576GB$ coherent memory and selected to perform measurements at very high resolutions even for memory-intensive methods.
Throughout the experiments, we set the block height $B=8$.
The fused kernel was implemented in \pname{NVIDIA CUTLASS CuTe} DSL,
targeted for
Ampere microarchitecture using warp-level MMA tensor cores with
\textit{bfloat16} inputs and \textit{float32} accumulators.

The correctness was verified with unit tests and observing the endpoint error
when used in \raft. It achieves near-zero $0.03\%$ endpoint error difference 
compared to the official implementation.

\subsection{Isolated Correlation Sampling Results}
\label{sec:results:isolated}

We measure the runtime and the peak memory consumption,
as reported by \textit{PyTorch},
by running each operation for all dataset image pairs and report the mean 
and standard deviation over all sample points.
As we observe only negligible variance between different runs with the same 
inputs, we measure a single run per image.

The default setting uses $2048 \times 896$ input image resolution, with the
correlation volume size of $(256 \times 112)^2$, $256$
feature channels and $32$ flow update iterations, matching the official \raft 
implementation.

\subsubsection{Image Resolution}
\label{sec:results:isolated:scale}

Primarily, we investigate the impact of the image resolution on the correlation
computation and report the results in Fig.~\ref{fig:benchmark_scale}.
As images are scaled uniformly, increasing the input width by $2\times$, increases the number of pixels by $4\times$
and the size of the dense correlation volume by $16\times$.

It can be observed that the default implementation has a quadratic memory increase and 
at $1024 \times 448$ resolution already requires $719GB$ to store the
dense correlation volume, becoming prohibitively large to compute.
On the other hand, \textit{on-demand} sampling maintains low memory usage
but significantly underperforms in runtime.

Our method achieves linear increase of both runtime
and memory in the number of input pixels.
Compared to prior methods, with fixed memory usage,
it achieves more than $90\%$ reduction in runtime, or,
at fixed runtime, it achieves up to $99\%$ reduction in memory usage.

The exact measurements are provided in the supplementary material.

\subsubsection{Other variables}

Additionally, we investigate the impact of the number of iterations, input feature 
dimensionality, as well as hardware, on the runtime and memory
of correlation sampling and provide results in the supplementary document.

Similar to the results described in Section~\ref{sec:results:isolated:scale}, they
show favorable runtime and memory trade-off, and all considered GPU models
show results consistent with those shown in Figure~\ref{fig:benchmark_scale}.
 
\setlength{\tabcolsep}{5pt}

\begin{table*}[h!]
    \small
    \centering
    \begin{tabular}{lcccccccc}
    \toprule   
   	& Best-Accuracy
   	& 1px error
   	& EPE
   	& LM - 1px error
   	& LM-EPE
   	& Best
   	& Without
   	& Our
\\
& Input Width
& $\%$ $\downarrow$
& px $\downarrow$
& $\%$ $\downarrow$
& px $\downarrow$
& Runtime, s
& Ours, s
& Improvement
    \small
\\\midrule
GMFlow~\cite{xuGMFlowLearningOptical2022} & {\color{Maroon}$1024$ - $\sfrac{1}{8}$} & $43.9$ & $2.95$ & $94.3$ & $24.46$ & $\bm{0.09}$ & \multicolumn{2}{c}{\texttt{\color{gray}n/a}}\\ 
PWC-Net~\cite{sun2018pwc} & {\color{Maroon}$4096$ - $\sfrac{1}{2}$} & $24.4$ & $3.26$ & $61.4$ & $57.14$ & $\underline{0.21}$ & \multicolumn{2}{c}{\texttt{\color{gray}n/a}}\\ 
Flow-1D~\cite{xuHighResolutionOpticalFlow2021} & {\color{Maroon}$4096$ - $\sfrac{1}{2}$} & $23.8$ & $2.23$ & $71.3$ & $31.58$ & $0.79$ & \multicolumn{2}{c}{\texttt{\color{gray}n/a}}\\ 
DIP~\cite{zhengDIPDeepInverse2022} & $8192$ - $\sfrac{1}{1}$ & $19.8$ & $4.00$ & $51.2$ & $85.62$ & $11.57$ & \multicolumn{2}{c}{\texttt{\color{gray}n/a}}\\ 
FlowFormer~\cite{huangFlowFormerTransformerArchitecture2022} & $8192$ - $\sfrac{1}{1}$ & $16.9$ & $3.22$ & $56.4$ & $58.16$ & $16.57$ & \multicolumn{2}{c}{\texttt{\color{gray}n/a}}\\ 
FlowFormer++~\cite{shiFlowFormerMaskedCost2023} & $8192$ - $\sfrac{1}{1}$ & $16.8$ & $3.41$ & $55.3$ & $55.70$ & $16.68$ & \multicolumn{2}{c}{\texttt{\color{gray}n/a}}\\ 
SCV~\cite{jiangLearningOpticalFlow2021} & {\color{Maroon}$4096$ - $\sfrac{1}{2}$} & $19.1$ & $2.83$ & $46.1$ & $46.42$ & $14.66$ & \multicolumn{2}{c}{\texttt{\color{gray}n/a}}\\ 
HCVFlow~\cite{zhaoHybridCostVolume2024} & {\color{Maroon}$4096$ - $\sfrac{1}{2}$} & $17.9$ & $2.08$ & $54.4$ & $32.92$ & $1.09$ & \multicolumn{2}{c}{\texttt{\color{gray}n/a}}\\ 
ReCoVEr-CX~\cite{Kiefhaber_2025_ICCV} & {\color{Maroon}$4096$ - $\sfrac{1}{2}$} & $22.6$ & $3.01$ & $94.6$ & $58.71$ & $0.56$ & \multicolumn{2}{c}{\texttt{\color{gray}n/a}}\\ 
WAFT~\cite{wangWAFTWarpingAloneField2025} & {\color{Maroon}$4096$ - $\sfrac{1}{2}$} & $15.3$ & $2.74$ & $62.6$ & $65.14$ & $8.47$ & \multicolumn{2}{c}{\texttt{\color{gray}n/a}}\\ 
\arrayrulecolor{black!50}\midrule
MS-RAFT+~\cite{jahediMSRAFTHighResolution2024} & {\color{Maroon}$4096$ - $\sfrac{1}{2}$} & $\underline{14.5}$ & $\underline{1.92}$ & $34.6$ & $32.03$ & $4.15$ & $5.51$ & {\color{OliveGreen}$-25\%$}\\ 
RAFT~\cite{teedRAFTRecurrentAllPairs2020} & {\color{Maroon}$4096$ - $\sfrac{1}{2}$} & $18.9$ & $5.90$ & $49.8$ & $36.41$ & $0.96$ & $2.62$ & {\color{OliveGreen}$-63\%$}\\ 
CCMR~\cite{jahediCCMRHighResolution2024} & {\color{Maroon}$4096$ - $\sfrac{1}{2}$} & $15.8$ & $2.16$ & $39.7$ & $42.86$ & $4.39$ & $5.23$ & {\color{OliveGreen}$-16\%$}\\ 
DPFlow~\cite{morimitsuDPFlowAdaptiveOptical2025} & $8192$ - $\sfrac{1}{1}$ & $16.5$ & $\underline{1.92}$ & $\underline{34.4}$ & $\bm{18.03}$ & $5.00$ & $5.48$ & {\color{OliveGreen}$-9\%$}\\ 
SEA-RAFT~\cite{wangSEARAFTSimpleEfficient2025} & {\color{Maroon}$4096$ - $\sfrac{1}{2}$} & $16.8$ & $4.94$ & $39.8$ & $39.83$ & $0.42$ & $0.62$ & {\color{OliveGreen}$-33\%$}\\ 
\multirow{2}{*}{SEA-RAFT (Cascaded)} & {\color{Maroon}$4096$ - $\sfrac{1}{2}$} & $15.8$ & $\bm{1.90}$ & $36.8$ & $\underline{18.58}$ & $0.45$ & $0.72$ & {\color{OliveGreen}$-37\%$}\\ 
 & $8192$ - $\sfrac{1}{1}$ & $\bm{13.3}$ & $2.70$ & $\bm{31.6}$ & $21.53$ & $1.91$ & $2.89$ & {\color{OliveGreen}$-34\%$}\\
\arrayrulecolor{black}\bottomrule
    \end{tabular}
    \vspace*{-1mm}
    \caption{
Quantitative evaluation of optical flow estimation methods on a $8K$ dataset.
For each method, we list the resolution that can be run with $80GB$ of GPU memory
and obtains the best 1px error.
For a full comparison with other methods, we also list our $\sfrac{1}{2}$ results.
We report the 1px outlier rate, endpoint-error (EPE), both metrics for pixels 
with large motion (LM, magnitude over 128px) and the best runtime across all 
variants with and without our improvements.
We highlight the best (in \textbf{bold}) and second-best (\underline{underlined}) method.
}
    \label{table:flow_eval}
    \vspace*{-1.5mm}
    \end{table*}
    
\setlength{\tabcolsep}{6pt}

\section{Ultra-High-Resolution Evaluation}
\label{sec:flow_evaluation}
    
    To perform zero-shot evaluation and benchmarking of optical flow estimation
    methods on UHR inputs, we render several sequences
    from the \textsc{Blender} movie \textsc{Charge} with rendered
    ground truth displacements.
    We then propose an inference extension
    to the \pname{SEA-RAFT} optical flow estimation method
    and perform extensive evaluation of the existing methods.
    We perform these tests with an \textit{NVIDIA A100 80GB} GPU.
    
    \subsection{Dataset}
    
    We follow the commonly used benchmarking
    approach~\cite{Mehl2023_Spring,Butler:ECCV:2012}  of generating
    frames and motion vectors from publicly accessible
    computer-generated movies.
We choose \textsc{Blender} movie \textsc{Charge}
    as a recent photo-realistic movie that is not 
    used in existing benchmarks.
    In total, we generate $335$ frames
    at $8192 \times 3432$px resolution with
    super-resolved rendered ground at $16K$ resolution,
    following Mehl~\etal~\cite{Mehl2023_Spring}.
Due to computational
    reasons, we only consider forward flow.
The dataset consists of $332$ evaluation pairs and
    tests for prediction of $9.3B$ pixels.
An example of from the dataset can be seen in
    Figure~\ref{fig:flow:qualitative}.
    
    Details on the dataset generation are provided in the supplementary material.
    
    \subsection{Cascaded Inference}
    
    Our initial experiments indicated that evaluation at
    high resolutions degrade the performance of estimating
    large displacements.
To mitigate it, we propose a simple cascaded
    test-time extension
    of \pname{SEA-RAFT} that requires no
    additional training.
    
    Before applying any iterative flow updates,
    we recursively initialize flows as a lower resolution
    estimate. More formally, whenever the minimum input 
    dimension is more than $800px$, we bilinearly downscale
    inputs to $\sfrac{1}{4}$ resolution, estimate the flow,
    and initialize the flows with $\sfrac{1}{2}$ downscaled
    outputs
    (note that flows are updated at $\sfrac{1}{8}$ resolution).
    
    It is akin to a multi-resolution version of the
    \textit{warm-start} initialization used in \raft, 
    and unlike \pname{MS-RAFT+}~\cite{jahediMSRAFTHighResolution2024}, it does not
    require training of multiple resolution modules.
    
\subsection{Evaluation Results}  
    
    We evaluate several methods on
    different levels of processing scale by bilinearly downsampling
    inputs to $\{\sfrac{1}{8}; \sfrac{1}{4}; \sfrac{1}{2}; \sfrac{1}{1}\}$
    of their resolution and bilinearly
    upscaling the flow outputs.
    
\paragraph{Flow accuracy.} Quantitative end-to-end evaluation results are reported in Table~\ref{table:flow_eval}.
For each method, we list the processing level of scale with the lowest 1 pixel
error across all scales the method could be run with $80GB$ of GPU memory,
and report the lowest runtime across feasible variants both with and
without our improvements.

First, we observe a significant reduction in runtime, achieving more than
$30\%$ improvement for the majority of \raft-based methods.
Second, our proposed cascaded inference achieves the best 1 pixel error, both overall
and for large-motion pixels. Third, the majority of prior methods either cannot
be run on the full 8K resolution inputs, or their quality degrades.
Overall, we show that with our improvements, cost-volume-based methods
can outperform prior efficient methods both in runtime and quality.

\begin{figure*}[t]
      \centering
      \includegraphics[width=0.97\linewidth]{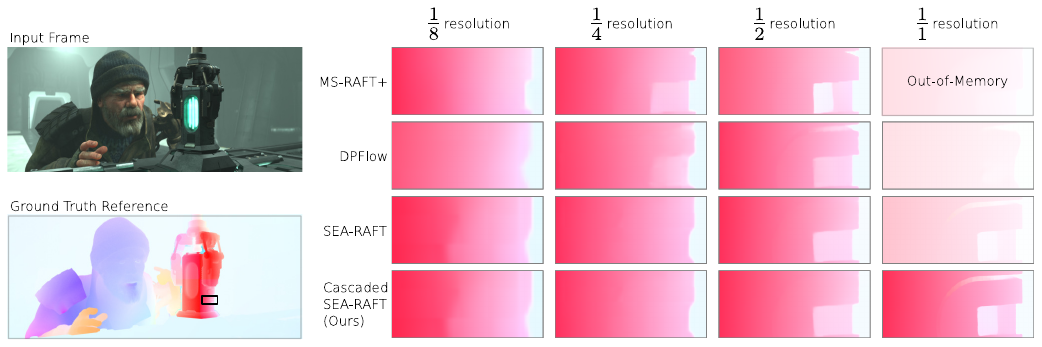}
      \vspace{-1mm}
      \caption{
      Qualitative comparison across different evaluation scales
      for
      MS-RAFT+~\cite{jahediMSRAFTHighResolution2024},
      DPFlow~\cite{morimitsuDPFlowAdaptiveOptical2025},
      and
      SEA-RAFT~\cite{wangSEARAFTSimpleEfficient2025}.
      {\scriptsize Image from Charge by Blender Studio.}
      }
      \label{fig:flow:qualitative}
\end{figure*}

Additionally, we investigate the impact of resolution on the prediction quality
and provide a qualitative comparison in Figure~\ref{fig:flow:qualitative},
showing that cascaded high-resolution inference allows estimating large motion
while retaining fine-grained details.
Full quantitative measurements are provided in the supplementary material.

\begin{table*}[]
\small
\centering
\begin{tabular}{lccccccc}
\toprule
& & \multicolumn{2}{c}{\bfseries Default}
& \multicolumn{2}{c}{\bfseries On-Demand Sampling}
& \multicolumn{2}{c}{\bfseries Ours}
\\
\cmidrule(lr){3-4} \cmidrule(lr){5-6} \cmidrule(lr){7-8}
Method & Input Width
 &  Runtime & Memory
 &  Runtime & Memory
 &  Runtime & Memory
\\
\midrule\parbox[t]{20mm}{\multirow{4}{*}{RAFT}} &$1024$ - $\sfrac{1}{8}$ &$0.07$ {\scriptsize ($-16\%$)}& $3.42$ {\scriptsize ({\color{Maroon}$+6\%$})} &$0.56$ {\scriptsize ({\color{Maroon}$+559\%$})}& $3.23$ {\scriptsize (=)} &$0.09$& $3.23$ \\ 
&$2048$ - $\sfrac{1}{4}$ &$0.25$ {\scriptsize ({\color{Maroon}$+2\%$})}& $8.78$ {\scriptsize ({\color{Maroon}$+115\%$})} &$0.89$ {\scriptsize ({\color{Maroon}$+261\%$})}& $4.08$ {\scriptsize (=)} &$0.25$& $4.08$ \\ 
&$4096$ - $\sfrac{1}{2}$ &\multicolumn{2}{c}{\texttt{\color{gray}OOM}} &$2.62$ {\scriptsize ({\color{Maroon}$+172\%$})}& $7.46$ {\scriptsize (=)} &$0.96$& $7.46$ \\ 
&$8192$ - $\sfrac{1}{1}$ &\multicolumn{2}{c}{\texttt{\color{gray}OOM}} &$10.47$ {\scriptsize ({\color{Maroon}$+206\%$})}& $20.96$ {\scriptsize (=)} &$3.43$& $20.96$ \\ 
\midrule 
\parbox[t]{20mm}{\multirow{3}{*}{MS-RAFT+}} &$1024$ - $\sfrac{1}{8}$ &\multicolumn{2}{c}{\texttt{\color{gray}OOM}} &$0.44$ {\scriptsize ({\color{Maroon}$+43\%$})}& $4.42$ {\scriptsize ($-1\%$)} &$0.31$& $4.45$ \\ 
&$2048$ - $\sfrac{1}{4}$ &\multicolumn{2}{c}{\texttt{\color{gray}OOM}} &$1.41$ {\scriptsize ({\color{Maroon}$+34\%$})}& $8.69$ {\scriptsize ($-1\%$)} &$1.05$& $8.80$ \\ 
&$4096$ - $\sfrac{1}{2}$ &\multicolumn{2}{c}{\texttt{\color{gray}OOM}} &$5.51$ {\scriptsize ({\color{Maroon}$+33\%$})}& $25.78$ {\scriptsize ($-2\%$)} &$4.15$& $26.21$ \\ 
\midrule 
\parbox[t]{20mm}{\multirow{4}{*}{DPFlow}} &$1024$ - $\sfrac{1}{8}$ &$0.14$ {\scriptsize ($-2\%$)}& $3.43$ {\scriptsize ({\color{Maroon}$+6\%$})} &$0.15$ {\scriptsize ({\color{Maroon}$+3\%$})}& $3.24$ {\scriptsize (=)} &$0.14$& $3.24$ \\ 
&$2048$ - $\sfrac{1}{4}$ &$0.40$ {\scriptsize ({\color{Maroon}$+12\%$})}& $9.53$ {\scriptsize ({\color{Maroon}$+146\%$})} &$0.41$ {\scriptsize ({\color{Maroon}$+15\%$})}& $3.87$ {\scriptsize (=)} &$0.35$& $3.87$ \\ 
&$4096$ - $\sfrac{1}{2}$ &\multicolumn{2}{c}{\texttt{\color{gray}OOM}} &$1.41$ {\scriptsize ({\color{Maroon}$+10\%$})}& $6.57$ {\scriptsize (=)} &$1.27$& $6.57$ \\ 
&$8192$ - $\sfrac{1}{1}$ &\multicolumn{2}{c}{\texttt{\color{gray}OOM}} &$5.48$ {\scriptsize ({\color{Maroon}$+10\%$})}& $17.38$ {\scriptsize (=)} &$5.00$& $17.38$ \\ 
\midrule 
\parbox[t]{20mm}{\multirow{4}{*}{SEA-RAFT}} &$1024$ - $\sfrac{1}{8}$ &$0.03$ {\scriptsize (=)}& $3.47$ {\scriptsize ({\color{Maroon}$+7\%$})} &$0.09$ {\scriptsize ({\color{Maroon}$+200\%$})}& $3.24$ {\scriptsize (=)} &$0.03$& $3.25$ \\ 
&$2048$ - $\sfrac{1}{4}$ &$0.13$ {\scriptsize ({\color{Maroon}$+28\%$})}& $8.86$ {\scriptsize ({\color{Maroon}$+146\%$})} &$0.18$ {\scriptsize ({\color{Maroon}$+76\%$})}& $3.56$ {\scriptsize ($-1\%$)} &$0.11$& $3.60$ \\ 
&$4096$ - $\sfrac{1}{2}$ &\multicolumn{2}{c}{\texttt{\color{gray}OOM}} &$0.62$ {\scriptsize ({\color{Maroon}$+49\%$})}& $5.22$ {\scriptsize (=)} &$0.42$& $5.22$ \\ 
&$8192$ - $\sfrac{1}{1}$ &\multicolumn{2}{c}{\texttt{\color{gray}OOM}} &$2.63$ {\scriptsize ({\color{Maroon}$+48\%$})}& $11.82$ {\scriptsize (=)} &$1.78$& $11.82$ \\ 
 \bottomrule
\end{tabular}
\vspace*{-1mm}
\caption{
Runtime (s) and peak memory usage (GiB) of the full
optical flow method end-to-end evaluation depending on the correlation computation 
variant at different scales of the inputs. We report the relative difference 
compared to our method in parentheses.
\texttt{OOM} indicates that the method requires more than $80GB$ of memory
and fails with an out-of-memory error.
}
\label{table:integration_results}
\end{table*}
 
\paragraph{Runtime and Memory.} In Table~\ref{table:integration_results}, we show the improvement in runtime and peak memory usage
of our method in an end-to-end
estimation, depending on the image resolution.

Across all methods and input resolutions, our method achieves similar or better runtime
as the default sampling with improvement in memory, and similar memory as the
\textit{on-demand} sampling with a significant improvement in runtime.

\paragraph{Additional Datasets.} As our performance improvements are agnostic to the input data,
we achieve comparable similar results on other optical flow
benchmarks, and provide results in the supplementary material.

\section{Conclusion}
\label{sec:conclusion}
In this paper,
we propose an efficient all-pairs correlation sampling algorithm,
which allows to significantly improve performance of {\raft}-based methods.
First, we analyze 
the existing approaches for all-pairs correlation volume sampling,
their volume sampling patterns in a practical optical flow application
and propose an algorithm that
utilizes these observations to perform correlation sampling more efficiently. 
In extensive experiments, we show that our method
achieves low memory consumption and runtime when compared to previous solutions.
Additionally, we evaluate existing methods on an \textit{8K} resolution
dataset and achieve state-of-the-art results on both in accuracy and performance.

In this work, we did not perform extensive low-level fine-tuning,
or architecture-specific optimizations for a fully optimized implementation.
Additionally, we only consider forward pass optimizations, as training is typically
done at lower resolutions due to memory requirements of other parts of the method.
However, the same optimizations can be applied to the backward pass, and we
leave it to future work to investigate that.

\FloatBarrier

{\small
\bibliographystyle{ieeenat_fullname}

}

\clearpage

\maketitlesupplementary

\section{Method Details}

In this section, we provide more information on obtaining the block mask,
the sampling patterns, as well as complexity analysis and
implementation details of our method.

\begin{figure*}[t]
      \centering
      \includegraphics[width=0.9\linewidth]{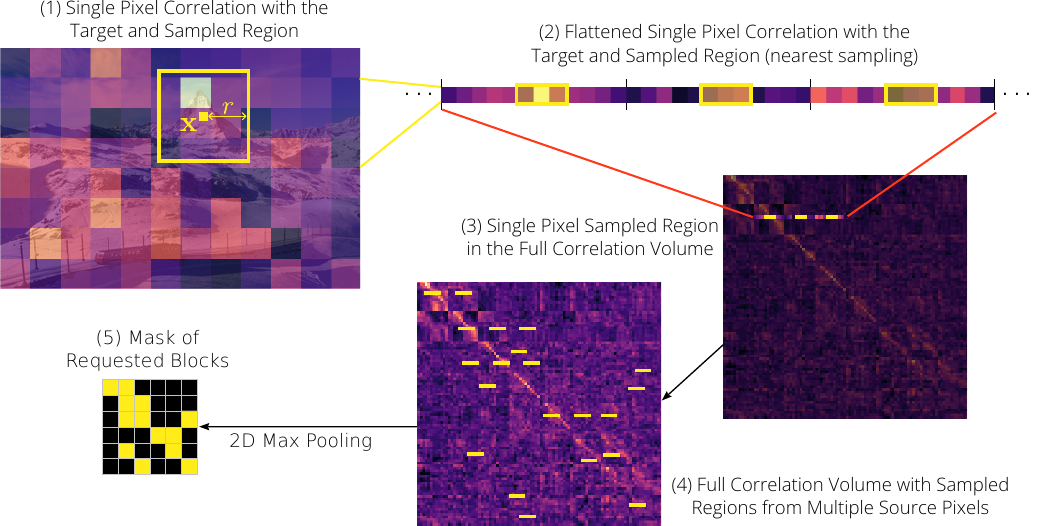}
      \vspace{-1mm}
      \caption{
        Conceptual visualization of the block mask computation.
      }
      \label{fig:block_mask}
\end{figure*}     

\subsection{Obtaining the Block Mask}

In Figure~\ref{fig:block_mask} we provide a conceptual visualization
of how the block mask is computed, which is then used for the correlation
volume analysis and to determine which blocks need to be computed.
Note that in practice it is obtained through splatting into the
block mask directly, or computed implicitly through the inter-thread
voting.

\subsection{Correlation Volume Sampling Patterns}

In Table~\ref{table:sparsity}, we show that
averaging the number of sampled values per row-major block
increases the ratio of cells
that need to be computed, but is still significantly lower than
the full matrix for sufficiently small block sizes.
In the second column, we show that by averaging rows in our patch-major manner,
sparsity is significantly increased without any computational overhead.
At the largest measured block size of $16^2$, it becomes too large to
maintain high level sparsity.

\begin{table}
  \small
  \centering
  \begin{tabular}{c|ccc}
  \toprule
  \textbf{Block Size} & \textbf{Row-Major} & \textbf{Patch-Major} & \textbf{Improvement}\\
  \midrule
  $1^2$  & \multicolumn{2}{c}{$1.6 \pm 0.4$} & $-$  \\
  $2^2$  & $2.5 \pm 0.5$  & $2.3 \pm 0.5$  & $8\%$ \\
  $4^2$  & $6.7 \pm 1.1$  & $4.2 \pm 0.7$  & $37\%$ \\
  $8^2$  & $20.6 \pm 4.4$ & $8.6 \pm 1.4$  & $58\%$ \\
  $16^2$ & $28.2 \pm 8.0$ & $27.1 \pm 1.2$ & $4\%$ \\
  \bottomrule
  \end{tabular}
  
  \caption{
  Percentage of sampled correlation volume cells depending on the block size and layout, measured over the \pname{SINTEL}~\cite{Butler:ECCV:2012} training dataset.
  }
  \label{table:sparsity}
  
\end{table}

\subsection{Implementation Details}

In this subsection, we provide more details on our implementation of the fused
correlation volume sampling algorithm, split in 3 parts according to our
algorithmic design:
1) computing implicit computation mask;
2) joint computation of the next correlation block;
3) correlation value sampling.
A simplified pseudocode is provided in \cref{alg:method},
kernel implementation is provided in Listings \ref{lst:kernel:a}-\ref{lst:kernel:c}.
It computes a single-level correlation whereas for multi-level pyramid
an average-pooled $F^{2}$ is used for each level.
We set $B=8$ as a trade-off between being able to
use efficient warp MMA operations for block computations
and limiting the shared memory usage.

\begin{algorithm*}[t]
\caption{Single-Level Correlation Volume Sampling}\label{alg:method}
\begin{algorithmic}
\Require Flattened input features $F^{1, 2} \in \mathbb{R}^{H \times W \times D}$, lookup centroids $\mathbf{X}\in \mathbb{R}^{N \times H \times W \times 2}$, block size $B$, lookup radius $r$, output buffer $\mathbf{O}\in \mathbb{R}^{N \times H \times W \times (2r + 1)^2}$.
\State 
\State $bH, bW \gets \ceil*{H / B}, \ceil*{W / B}$ \Comment{Setting the number of blocks}
\State $\bar{F}^{1, 2} \gets$ rearrange($\bar{F}^{1, 2}$, '$(bH{\times}B_h){\times}(bW{\times}B_w){\times}D \rightarrow (bH{\times}bW){\times}(B_h{\times}B_w){\times}D$') \Comment{Patch-major reshaping}
\State $\texttt{nBlocks} \gets bH \cdot bW$ \Comment{The total number of target blocks}
\For{$i = 0, 1, \dots, N-1$} \Comment{For every lookup iteration}
\CudaKernel{block $by \in [bH \cdot bW]$, thread $t \in [B^2]$}
    \State $\mathbf{x} \gets \mathbf{X}[i][by \cdot B^2 + t]$  \Comment{Get the respective target coordinates}
    \State $\texttt{blocks} \gets [\quad]$ \Comment{Empty list of blocks to compute}
    \ForAll{$\mathbf{dx'} \in \{-r, -r+1, ..., r, r+1\}^2$} \Comment{Find blocks for all offsets}
        \State $\texttt{blockId} \gets \floor*{(\mathbf{x} + \mathbf{dx}) / B}$
        \If{$0 <= \texttt{blockId} < \{bH, bW\} \texttt{ and (empty(blocks) or blockId} > \texttt{blocks[-1]}$)}
            \State $\texttt{blocks.append(blockId)}$
        \EndIf
    \EndFor
    \State $\texttt{blocks.append(nBlocks)}$ \Comment{Add end marker}
    \State
    \While{True}
        \State $\texttt{nextBlock} \gets $ inter-thread minimum of $\texttt{blocks.front()}$ \Comment{Vote for the next block to compute}
        \If{\texttt{nextBlock >= nBlocks}}
            \State \textbf{break} \Comment{All blocks have been processed}
        \EndIf
        \State 
        \State $\texttt{corrTile} \in \mathbb{R}^{B^2 \times B^2} \gets (\bar{F}^1[by])(\bar{F}^2[\texttt{nextBlock}])^T$ \Comment{Matrix-matrix multiplication of the block}
        \State 
        \If{\texttt{nextBlock = blocks.front()}} \Comment{This thread requested the current block}
            \State \texttt{blocks.pop(0)} \Comment{Advance local block pointer}
            \ForAll{$\mathbf{dx} \in \{-r, -r+1, ..., r\}^2$} \Comment{Sample from the computed block}
                \State $\mathbf{x{'}} \gets \mathbf{x} + \mathbf{dx} - \texttt{nextBlock} \cdot B$ \Comment{Local coordinates within the block} 
                \State $\mathbf{O}[i][by \cdot B^2 + t][\mathbf{dx}] \gets \textbf{sample} \texttt{ corrTile}[t][\mathbf{x{'}}]$
            \EndFor
        \EndIf
    \EndWhile
\EndCudaKernel
\EndFor
\end{algorithmic}
\end{algorithm*}
 
\paragraph{Computing Implicit Block Mask.} To determine the blocks which need to be computed, in each thread we iterate
over the $[-r, r]^2$ neighborhood around the target pixel and compute
indices of blocks the value needs to be sampled from.
In practice, it is not necessary to iterate over all offsets, as for any 3 consecutive
offsets, at least 2 land in the same block. Therefore, we only iterate over offsets with
a stride of the block size $B$.

Indices of all blocks that a local thread needs to compute are stored in a register array
in strictly increasing order, and we store an index pointer to the first block.
To determine the next block to compute across all threads in a thread block,
we perform a parallel reduction of the local indices
with atomic minimum operation to a shared memory
scalar, initialized larger than the size of the row.
If the current local block matches the next global block, the index pointer is advanced.
If the global next block is larger than the total number of blocks,
all blocks have been processed and the kernel can exit.

\paragraph{Computation of the Next Block.} Having agreed on the next block that needs to be computed, all threads participate
in its computation, distributing loading of the input values and performing
matrix-matrix computations across all threads.
At once, we perform \texttt{64x64x32} product with \texttt{16x8x16} MMA instructions,
\texttt{bf16} inputs and \texttt{fp32} accumulation. 
The final correlation values are kept in registers for faster processing.

\paragraph{Processing of the Correlation Block.} Only threads that voted for the currently computed block sample correlation
values by computing the target coordinates for each offset, and splitting them
into the block index and local coordinates within the block.
Values that can be extracted from the block are added to a global final output
buffer, initialized as zeros.

\subsection{Component Analysis}

\bgroup
\begin{table}[t]
    \small
\begin{tabular}{lrrr}
\toprule
\multicolumn{1}{r}{\textit{Correlation Volume Width =}}
& {\bfseries \boldmath $256$px}
& {\bfseries \boldmath $512$px}
\\
\midrule
{Preprocessing}& $11.3\%$& $9.5\%$\\ 
\addlinespace[2pt]
Computing Per-Thread Blocks& $\bm{32.7\%}$& $2.6\%$\\ 
\addlinespace[2pt]
Next-Block Voting& $0.2\%$& $0.3\%$\\ 
\addlinespace[2pt]
Matrix-Matrix Multiplication& $0.4\%$& $\bm{25.5\%}$\\ 
\addlinespace[2pt]
Sampling Blocks and Writing Outputs& $\bm{55.3\%}$& $\bm{62.2\%}$\\ 
\addlinespace[2pt]
\midrule
Total Runtime, ms& $25.9$& $93.9$\\ 
\addlinespace[2pt]
        \bottomrule
    \end{tabular}
\vspace*{-2mm}
\centering
\caption{
Approximate runtime breakdown of the steps of our method.
}
\label{table:component_results}
\vspace*{-2mm}       
    
\end{table}
\egroup 
To analyze the impact of different steps of our algorithm to the total runtime,
we approximate the time spent in each step by adding early kernel exits,
and observe the aggregated runtime increase over the whole dataset.
The runtime breakdown is provided in~\cref{table:component_results}.

\subsection{Memory Complexity Analysis}

Here we show that our algorithm achieves linear time and space complexity
in the number of pixels.

Assuming equal-size feature maps, let $P = H \times W$ be the number of
feature pixels per image, $B$ the block size, and $K=2r+1$ the lookup region size.
We also assume that both input tiles are fully stored in shared memory.
In practice, larger channel sizes $D$ is split into chunks of 32 elements.

As all lookup offsets are continuous, each source pixel can sample from $(K + 1)^2$
pixels. In the worst case, the first pixel is in a separate block, and, in
total, they spread across $(1 + \ceil*{\frac{(K + 1) - 1}{B}})^2=\floor*{\frac{K + 2B - 1}{B}}^2$ blocks.
With our hyperparameters $r=4, B=8$, it results in at most $3^2=9$ blocks per pixel.

The matrix-matrix multiplication for computing each tile requires $B^4D$ operations,
sampling it requires $\min(B^2, K^2)$ operations (if the block is smaller than the
lookup region, at most all values are sampled).
The total time complexity across all steps of the algorithm is therefore:
\begin{align}
    \label{eq:sup:time_complexity}
    P &\floor*{\frac{K + 2B - 1}{B}}^2 \Big(
    \underbrace{1}_{\text{Obtain mask}}
    + \underbrace{B^4D}_{\text{Compute a tile}} 
    + \underbrace{\min(B^2, K^2)}_{\text{Sample correlation}}
  \Big) \nonumber\\
\leq & P (K + 2B - 1)^2\frac{1}{B^2}(1 + B^4D + \min(B^2, K^2)) \nonumber\\
= & P (K + 2B - 1)^2(\frac{1}{B^2} + B^2D + \min(1, \frac{K^2}{B^2})) \\
\in & O(P (K + B)^2 B^2D),\nonumber
\end{align}
which is linear in the number of input pixels and feature
dimensionality, and quadratic in the lookup radius.
In practice, the number of requested blocks is lower due to smoothness of the optical flow and multiple source pixels requesting the same block, as shown in Table~\ref{table:sparsity}.

In addition to storing the inputs $O(PD)$ and outputs $O(PK^2)$,
we also need to use registers and shared memory to store the block indices and
inputs and outputs of one computed tile per thread block,
requiring:
\begin{align}
    \label{eq:sup:space_complexity}
    & P\floor*{\frac{K + 2B - 1}{B}}^2 + \frac{P}{B^2} \Big(
        2 B^2D + B^4)\nonumber\\
\leq & P ((\frac{K - 1}{B^2})^2+2) + P(2D + B^2) \\
\in & O(P (K^2 + D + B^2)),\nonumber
\end{align}
which, as inputs, is also linear in the number of pixels and feature dimensionality,
quadratic in the lookup radius.

\section{Extended Evaluation}

In this section we provide extended evaluation results.

\subsection{Correlation Sampling Measurements}

In Table~\ref{table:corr_benchmark}, we provide full benchmarking
results, corresponding to all runtime and memory usage plots.
We further analyze the impact of the number of iterations, input feature dimensions, and GPU model.
By default, $(512 \times 224)^2$ correlation volume, $256$
feature channels, and $32$ flow update iterations are used.

\begin{figure*}[t]
    \centering
    \begin{minipage}[b]{0.47\linewidth}
        \centering
        \includegraphics[width=\linewidth]{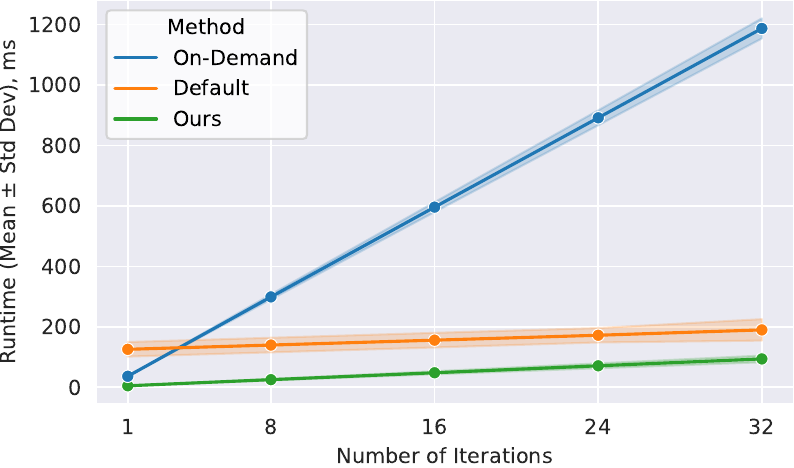}
    \end{minipage}\hfill
    \begin{minipage}[b]{0.47\linewidth}
        \centering
        \includegraphics[width=\linewidth]{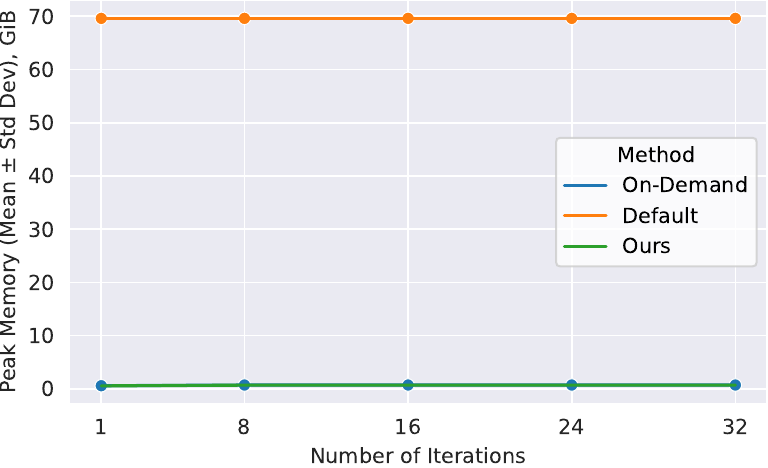}
    \end{minipage}
    \vspace{-2mm}
    \caption{
    Runtime and peak memory consumption by the number of flow update iterations.
    }
    \label{fig:benchmark_iterations}
\end{figure*}

\subsubsection{Number of Iterations}
In Figure~\ref{fig:benchmark_iterations}, we show the impact of number of flow update
iterations on the runtime and memory usage.
All methods show approximately linear runtime increase and 
constant peak memory consumption.
The gap between our method and the default implementation slightly narrows
with increasing number of iterations, as it has a fixed correlation volume
precomputation cost, while in further steps only sampling is performed.

\begin{figure*}[t]
    \centering
    \begin{minipage}[b]{0.47\linewidth}
        \centering
        \includegraphics[width=\linewidth]{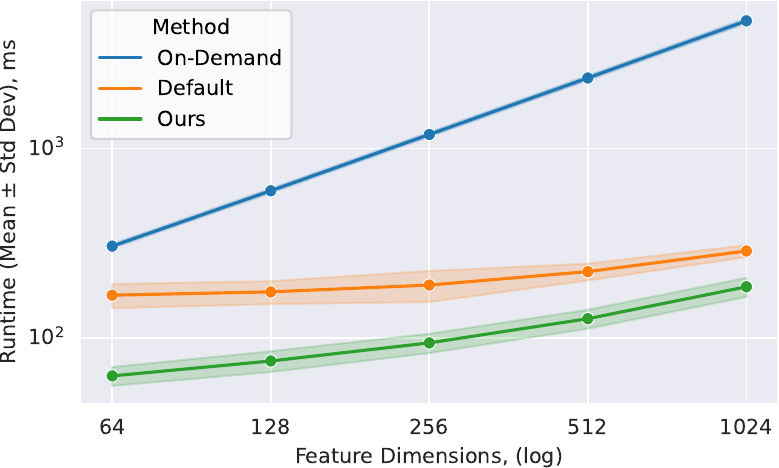}
    \end{minipage}\hfill
    \begin{minipage}[b]{0.47\linewidth}
        \centering
        \includegraphics[width=\linewidth]{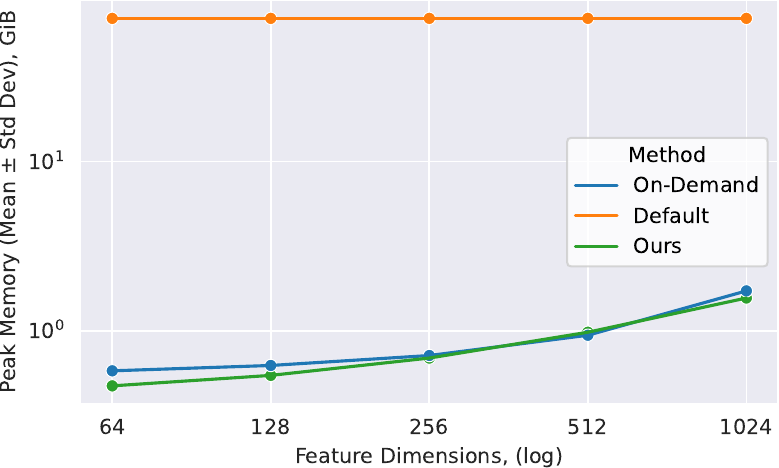}
    \end{minipage}
    \vspace{-2mm}    
    \caption{Runtime and peak memory consumption by input feature dimensionality. }
    \label{fig:benchmark_channels}
\end{figure*}

\subsubsection{Feature Dimensionality}
In Figure~\ref{fig:benchmark_channels}, we report the results at different feature
$F^{1, 2}$ dimensionality. 
The runtime of all methods increases approximately linearly with the feature dimension,
with slower rate for our method and the default implementation. 

As both our method and on-demand sampling stores down-sampled
replicas of the target features for multi-level computations,
they show an increase of memory with larger dimensions.
However, the absolute increase is small and typically negligible.

\subsubsection{Hardware}

\begin{figure*}[t]
    \centering
    \includegraphics[width=0.9\linewidth]{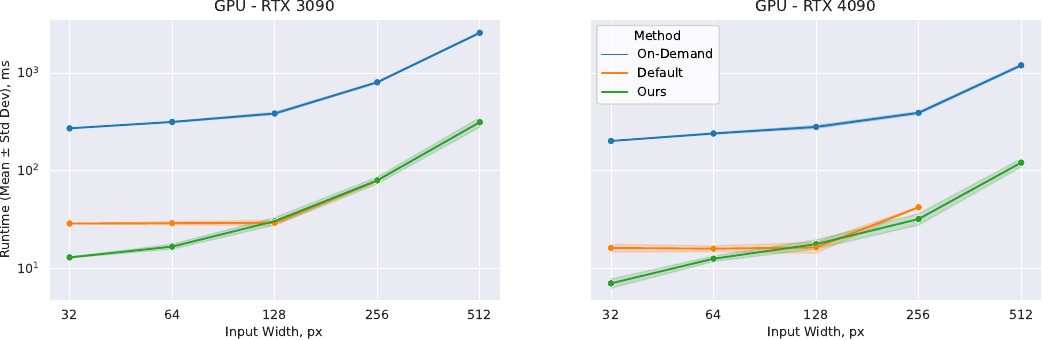}\\
    \vspace{+2mm}
    \includegraphics[width=0.9\linewidth]{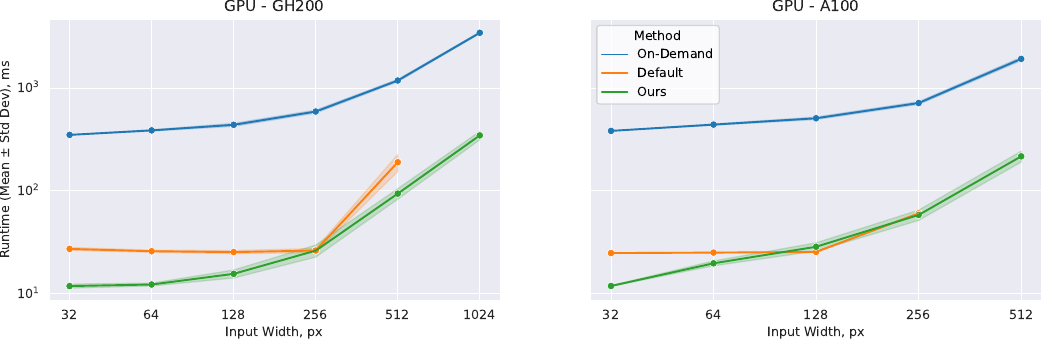}
    \caption{
    Runtime depending on input image size and different hardware.
    }
    \label{fig:benchmark_gpu}
\end{figure*}

We perform additional experiments with different
GPU models -
\textit{NVIDIA RTX 3090}, \textit{NVIDIA RTX 4090}, and \textit{NVIDIA A100}
- and show results in Figure~\ref{fig:benchmark_gpu}.
It can be observed that hardware has little impact on the relative
performance between different methods.

Note that for some GPU models, such as the \textit{A100},
at $128px$ correlation volume width, the overhead of pre-computing the full
cost matrix becomes less significant, slightly outperforming our method.
However, it still obtained at the expense of much higher memory requirements.

\subsection{Dataset Generation Details}

\begin{table}
  \small
  \centering
  \begin{tabular}{ccc}
  \toprule
  \textbf{Shot} & \textbf{Start Frame} & \textbf{End Frame}\\
  \midrule
  \textsc{010\_0050} & $101$ & $196$  \\
  \textsc{040\_0040} & $101$ & $264$  \\
  \textsc{060\_0130} & $101$ & $175$  \\
  \bottomrule
  \end{tabular}
  \vspace*{-2mm}
  \caption{
  List of rendered \textsc{Charge} sequences.
  }
  \vspace*{-3mm}
  \label{table:blender_charge_seq}
  
\end{table} 
For the data generation, we use the \textsc{Cycles} renderer to 
generate $335$ frames from $3$ sequences
of the \textsc{Blender} movie \textsc{Charge}
as listed in Table~\ref{table:blender_charge_seq}.
Sample frames from each of the sequences can be seen in Figures~\ref{fig:flow:compare_010}-\ref{fig:flow:compare_060}.

To improve the quality of the rendered motion vectors, we
remove lens flares and volumes, and disable motion blur to
decrease noise levels.
Following Mehl~\etal~\cite{Mehl2023_Spring},
we render super-resolved motion vectors at $16K$ resolution with four ground truth values
for each pixel.
The \pname{Python} script, used to apply scene adjustments,
is provided in Listing~\ref{lst:blender}.

\subsection{Extended Results}

In Table~\ref{tab:supplementary:mixed_strategies}, we further analyze the
improvements from our method by considering alternative strategies to
reduce computational costs.
First, we consider \pname{SEA-RAFT} cascaded variants with more
iterations at the lower resolution, but observe significant reduction
in flow quality.
Second, instead of using our correlation sampler, we run the default
implementation on internal processing resolutions it is feasible on 80GB GPU,
before switching to the on-demand sampler.
While small improvement over on-demand sampler is achieved, it is significantly
outperformed by our algorithm. 

In Table~\ref{table:full_results}, we provide the full results of all evaluated methods
on the \pname{Charge} $8K$ dataset.
The first part of the table contains methods that do not employ correlation volume sampling
and do not benefit from performance improvements using our efficient
volume sampling algorithm.

In Figure~\ref{fig:sup:lines_charge}, we visually plot the 1px error results
and runtime depending on the input scaling resolution. It can be observed that several
method degrade in quality when used with the highest resolution inputs.

\begin{figure*}
  \centering
  \noindent

\begin{subfigure}{\linewidth}
  \begin{minipage}{\linewidth}
    \centering
    \resizebox{\linewidth}{!}{\input{include/keoiutiasbgzzxlkfwnt.pgf}
    }
  \end{minipage}
  \caption{
  Comparison of 1px error.
  }
\end{subfigure}

\bigskip

\begin{subfigure}{\linewidth}
  \begin{minipage}{\linewidth}
    \centering
    \resizebox{\linewidth}{!}{\input{include/wwnrdkgqqmewdwffuhre.pgf}
    }
  \end{minipage}
  \caption{
    Comparison of runtime, normalized by the scaling factor.
  }
\end{subfigure}

\caption{
    Quality and runtime comparison of selected methods on the \textsc{Charge}
    dataset depending on the input scaling resolution.
}
    \label{fig:sup:lines_charge}
\end{figure*}

\begin{figure*}[p!]
\centering
\includegraphics[width=0.96\linewidth]{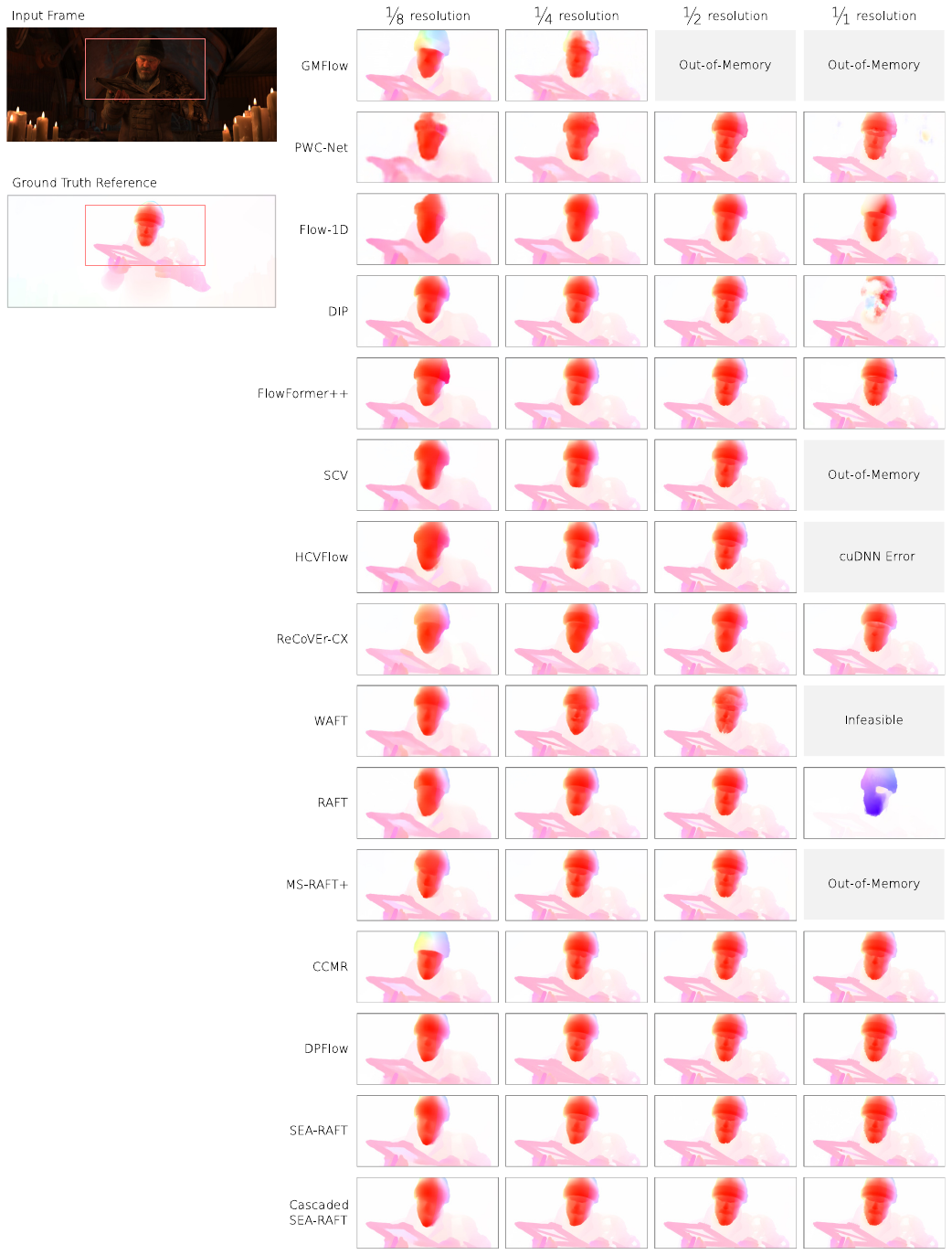}
\vspace*{-2.9mm}
\caption{
Qualitative comparison across different evaluation scales
on a frame from \texttt{010\_0050} shot.
{\scriptsize Image from Charge by Blender Studio.}
}
\label{fig:flow:compare_010}
\end{figure*}

\begin{figure*}[p!]
\centering
\includegraphics[width=0.96\linewidth]{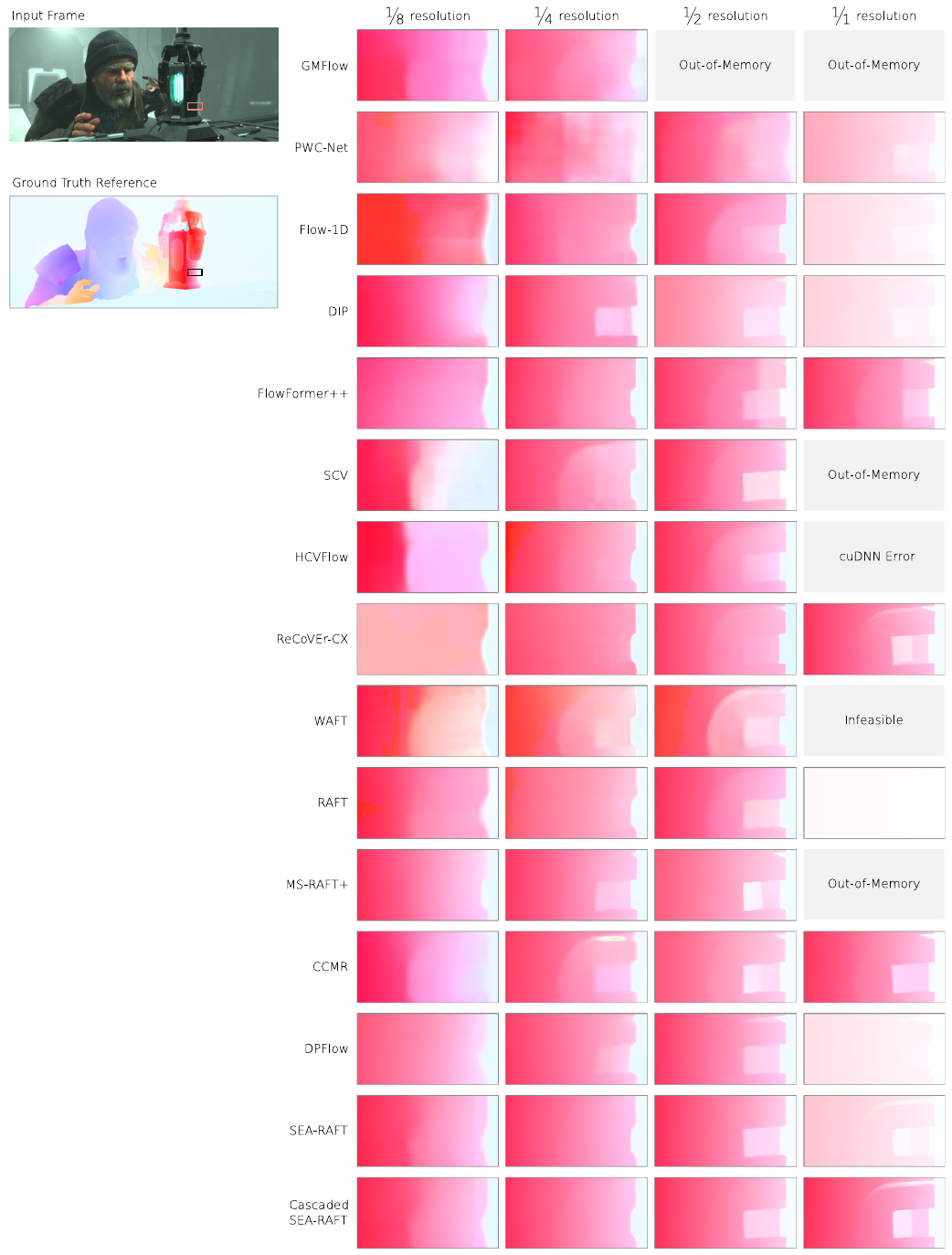}
\vspace*{-2.9mm}
\caption{
Qualitative comparison across different evaluation scales
on a frame from \texttt{040\_0040} shot.
{\scriptsize Image from Charge by Blender Studio.}
}
\label{fig:flow:compare_040}
\end{figure*}

\begin{figure*}[p!]
\centering
\includegraphics[width=0.96\linewidth]{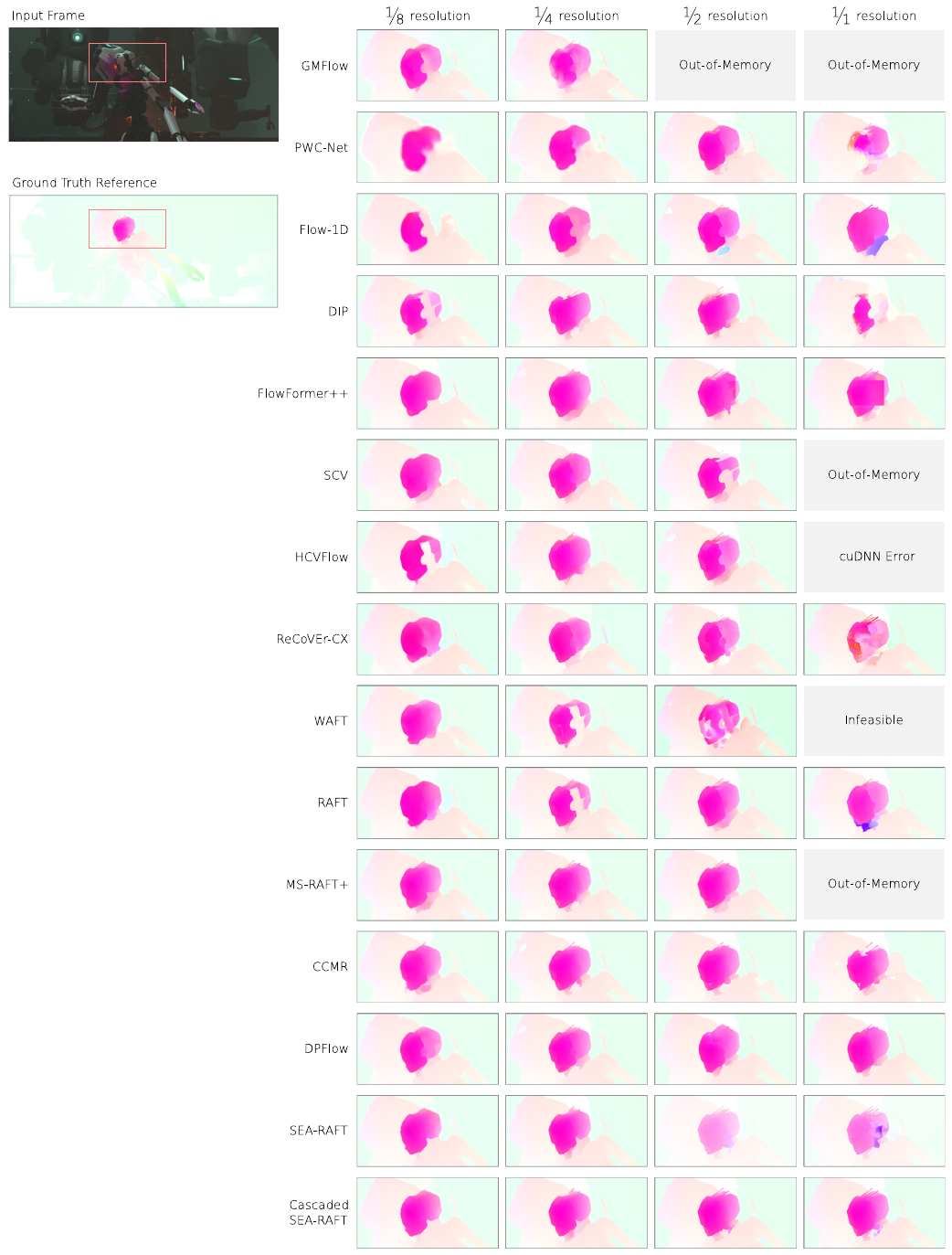}
\vspace*{-2.9mm}
\caption{
Qualitative comparison across different evaluation scales
on a frame from \texttt{060\_0130} shot.
{\scriptsize Image from Charge by Blender Studio.}
}
\label{fig:flow:compare_060}
\end{figure*}

In Figures~\ref{fig:flow:compare_010}-\ref{fig:flow:compare_060}, we provide additional qualitative comparisons.

To obtain results of all methods, we extend PTLFlow~\cite{morimitsu2021ptlflow},
and use checkpoints 
fined-tuned on the Sintel dataset, unless otherwise noted in Table~\ref{table:full_results}.
\pname{FlowFormer} results were obtained with their default tiling technique.
We do not run \pname{WAFT} on the full resolution due to its very slow runtime
($\approx2$ min / image).

\subsection{Additional Datasets}

We run similar evaluations on the \dname{Kubric-8K}~\cite{morimitsuDPFlowAdaptiveOptical2025}
dataset, which also contains ultra-high-resolution synthetic ground truth, but is
obtained from generated scenes with moving objects.
The respective results are provided in Figure~\ref{fig:sup:lines_kubric} and
Tables \ref{table:supl:flow_eval:kubric} and \ref{table:sup:kubric_raft_integration_results}.
Unlike the \pname{Charge} dataset, where several methods benefit from
fine-grained detail reconstruction at the highest resolution,
on \dname{Kubric-8K} most methods reach their lowest 1px error at quarter
resolution.
This could be explained from the use of less-detailed objects in the dataset.
However, our algorithm still provides runtime and memory improvements across
different evaluations scales (Table~\ref{table:sup:kubric_raft_integration_results}).

In addition to the \textit{8K} datasets used in most experiments, we
perform end-to-end runtime measurements on
\dname{Spring}~\cite{Mehl2023_Spring},
\dname{Sintel}~\cite{Butler:ECCV:2012} and
\dname{KITTI}~\cite{menzeObjectSceneFlow2015} datasets
and report results in
Table~\ref{table:sup:integration_additional_datasets}.
Despite these datasets having a much lower resolution than our target,
making the fast default implementation feasible,
our method can still provide significant runtime or memory improvements.
As the cascaded inference is not applicable on low-resolution inputs of these datasets,
and we focus on performance improvements, we
refer to their official benchmarks for full numerical results.
 
\begin{figure*}
  \centering
  \noindent

\begin{subfigure}{\linewidth}
  \begin{minipage}{\linewidth}
    \centering
    \resizebox{\linewidth}{!}{\input{include/cyocnvkmjgoxrotjrrvm.pgf}
    }
  \end{minipage}
  \caption{
  Comparison of 1px error.
  }
\end{subfigure}

\bigskip

\begin{subfigure}{\linewidth}
  \begin{minipage}{\linewidth}
    \centering
    \resizebox{\linewidth}{!}{\input{include/ldjxylioeudhncsvluqi.pgf}
    }
  \end{minipage}
  \caption{
    Comparison of runtime, normalized by the scaling factor.
  }
\end{subfigure}

\caption{
    Quality and runtime comparison of selected methods on the \textbf{\dname{Kubric-8K}}
    dataset depending on the input scaling resolution.
}
\label{fig:sup:lines_kubric}
\end{figure*}
 \setlength{\tabcolsep}{5pt}

\begin{table*}[h!]
    \small
    \centering
    \begin{tabular}{lcccccccc}
    \toprule   
   	& Best-Accuracy
   	& 1px error
   	& EPE
   	& LM - 1px error
   	& LM-EPE
   	& Best
   	& Without
   	& Our
\\
& Input Width
& $\%$ $\downarrow$
& px $\downarrow$
& $\%$ $\downarrow$
& px $\downarrow$
& Runtime, s
& Ours, s
& Improvement
    \small
\\\midrule
GMFlow~\cite{xuGMFlowLearningOptical2022} & {\color{Maroon}$2048$ - $\sfrac{1}{4}$} & $73.4$ & $5.08$ & $84.4$ & $15.36$ & $0.94$ & \multicolumn{2}{c}{\texttt{\color{gray}n/a}}\\ 
PWC-Net~\cite{sun2018pwc} & {\color{Maroon}$4096$ - $\sfrac{1}{2}$} & $19.7$ & $21.79$ & $44.5$ & $98.94$ & $0.22$ & \multicolumn{2}{c}{\texttt{\color{gray}n/a}}\\ 
Flow-1D~\cite{xuHighResolutionOpticalFlow2021} & {\color{Maroon}$4096$ - $\sfrac{1}{2}$} & $24.3$ & $7.80$ & $49.3$ & $31.45$ & $0.93$ & \multicolumn{2}{c}{\texttt{\color{gray}n/a}}\\ 
DIP~\cite{zhengDIPDeepInverse2022} & {\color{Maroon}$2048$ - $\sfrac{1}{4}$} & $14.9$ & $6.44$ & $33.5$ & $27.09$ & $0.96$ & \multicolumn{2}{c}{\texttt{\color{gray}n/a}}\\ 
FlowFormer~\cite{huangFlowFormerTransformerArchitecture2022} & {\color{Maroon}$2048$ - $\sfrac{1}{4}$} & $15.6$ & $4.10$ & $33.7$ & $15.92$ & $1.79$ & \multicolumn{2}{c}{\texttt{\color{gray}n/a}}\\ 
FlowFormer++~\cite{shiFlowFormerMaskedCost2023} & {\color{Maroon}$2048$ - $\sfrac{1}{4}$} & $14.7$ & $4.06$ & $32.5$ & $14.87$ & $1.80$ & \multicolumn{2}{c}{\texttt{\color{gray}n/a}}\\ 
SCV~\cite{jiangLearningOpticalFlow2021} & {\color{Maroon}$2048$ - $\sfrac{1}{4}$} & $22.8$ & $4.41$ & $39.8$ & $17.29$ & $3.48$ & \multicolumn{2}{c}{\texttt{\color{gray}n/a}}\\ 
HCVFlow~\cite{zhaoHybridCostVolume2024} & {\color{Maroon}$2048$ - $\sfrac{1}{4}$} & $15.3$ & $4.34$ & $36.2$ & $17.61$ & $0.34$ & \multicolumn{2}{c}{\texttt{\color{gray}n/a}}\\ 
ReCoVEr-CX~\cite{Kiefhaber_2025_ICCV} & {\color{Maroon}$4096$ - $\sfrac{1}{2}$} & $33.3$ & $20.33$ & $83.2$ & $93.02$ & $0.66$ & \multicolumn{2}{c}{\texttt{\color{gray}n/a}}\\ 
WAFT~\cite{wangWAFTWarpingAloneField2025} & {\color{Maroon}$2048$ - $\sfrac{1}{4}$} & $11.4$ & $4.81$ & $30.8$ & $20.17$ & $1.01$ & \multicolumn{2}{c}{\texttt{\color{gray}n/a}}\\ 
\arrayrulecolor{black!50}\midrule
MS-RAFT+~\cite{jahediMSRAFTHighResolution2024} & {\color{Maroon}$2048$ - $\sfrac{1}{4}$} & $10.9$ & $4.46$ & $22.0$ & $18.95$ & $1.23$ & $1.65$ & {\color{OliveGreen}$-26\%$}\\ 
RAFT~\cite{teedRAFTRecurrentAllPairs2020} & {\color{Maroon}$2048$ - $\sfrac{1}{4}$} & $17.7$ & $4.81$ & $38.8$ & $19.73$ & $0.30$ & $0.31$ & {\color{OliveGreen}$-4\%$}\\ 
CCMR~\cite{jahediCCMRHighResolution2024} & {\color{Maroon}$2048$ - $\sfrac{1}{4}$} & $13.5$ & $4.79$ & $26.2$ & $20.07$ & $1.29$ & $1.60$ & {\color{OliveGreen}$-20\%$}\\ 
DPFlow~\cite{morimitsuDPFlowAdaptiveOptical2025} & {\color{Maroon}$2048$ - $\sfrac{1}{4}$} & $\bm{8.0}$ & $\underline{3.16}$ & $\bm{21.0}$ & $\bm{11.22}$ & $0.38$ & $0.42$ & {\color{OliveGreen}$-11\%$}\\ 
SEA-RAFT~\cite{wangSEARAFTSimpleEfficient2025} & {\color{Maroon}$2048$ - $\sfrac{1}{4}$} & $10.2$ & $4.25$ & $23.7$ & $17.92$ & $\bm{0.12}$ & $0.17$ & {\color{OliveGreen}$-25\%$}\\ 
SEA-RAFT (Cascaded) 1/2 & {\color{Maroon}$4096$ - $\sfrac{1}{2}$} & $11.8$ & $3.95$ & $\underline{21.3}$ & $14.75$ & $0.53$ & $0.87$ & {\color{OliveGreen}$-39\%$}\\ 
SEA-RAFT (Cascaded) & {\color{Maroon}$2048$ - $\sfrac{1}{4}$} & $\underline{10.0}$ & $\bm{3.04}$ & $22.5$ & $\underline{11.78}$ & $\underline{0.14}$ & $0.18$ & {\color{OliveGreen}$-22\%$}\\ 
\arrayrulecolor{black}\bottomrule
    \end{tabular}
    \vspace*{-1mm}
    \caption{
Quantitative evaluation of optical flow estimation methods on the \textbf{\dname{Kubric-8K}} dataset.
For each method, we list the resolution that can be run with $80GB$ of GPU memory
and obtains the best 1px error.
For a full comparison with other methods we also list our $\sfrac{1}{2}$ results.
We report the 1px outlier rate, endpoint-error (EPE), both metrics for pixels 
with large motion (LM, magnitude over 128px) and the best runtime across all 
variants with and without our improvements.
We highlight the best (in \textbf{bold}) and second-best (\underline{underlined}) method.
    }
    \label{table:supl:flow_eval:kubric}
    \vspace*{-1.5mm}
    \end{table*}
    
\setlength{\tabcolsep}{6pt}    
 \begin{table*}[]
\small
\centering
\begin{tabular}{lccccccc}
\toprule
& & \multicolumn{2}{c}{\bfseries Default}
& \multicolumn{2}{c}{\bfseries On-Demand Sampling}
& \multicolumn{2}{c}{\bfseries Ours}
\\
\cmidrule(lr){3-4} \cmidrule(lr){5-6} \cmidrule(lr){7-8}
Method & Input Width
 &  Runtime & Memory
 &  Runtime & Memory
 &  Runtime & Memory
\\
\midrule\parbox[t]{20mm}{\multirow{4}{*}{RAFT}} &$960$ - $\sfrac{1}{8}$ &$0.09$ {\scriptsize ($-13\%$)}& $2.54$ {\scriptsize ({\color{Maroon}$+9\%$})} &$0.61$ {\scriptsize ({\color{Maroon}$+510\%$})}& $2.33$ {\scriptsize (=)} &$0.10$& $2.33$ \\ 
 &$1920$ - $\sfrac{1}{4}$ &$0.31$ {\scriptsize ({\color{Maroon}$+4\%$})}& $9.98$ {\scriptsize ({\color{Maroon}$+200\%$})} &$0.99$ {\scriptsize ({\color{Maroon}$+235\%$})}& $3.33$ {\scriptsize (=)} &$0.30$& $3.33$ \\ 
 &$3840$ - $\sfrac{1}{2}$ &\multicolumn{2}{c}{\texttt{\color{gray}OOM}} &$3.34$ {\scriptsize ({\color{Maroon}$+197\%$})}& $7.31$ {\scriptsize (=)} &$1.12$& $7.31$ \\ 
 &$7680$ - $\sfrac{1}{1}$ &\multicolumn{2}{c}{\texttt{\color{gray}OOM}} &$12.49$ {\scriptsize ({\color{Maroon}$+212\%$})}& $23.26$ {\scriptsize (=)} &$4.00$& $23.26$ \\ 
 \midrule 
\parbox[t]{20mm}{\multirow{3}{*}{MS-RAFT+}} &$960$ - $\sfrac{1}{8}$ &\multicolumn{2}{c}{\texttt{\color{gray}OOM}} &$0.52$ {\scriptsize ({\color{Maroon}$+43\%$})}& $3.73$ {\scriptsize ($-1\%$)} &$0.36$& $3.76$ \\ 
 &$1920$ - $\sfrac{1}{4}$ &\multicolumn{2}{c}{\texttt{\color{gray}OOM}} &$1.65$ {\scriptsize ({\color{Maroon}$+34\%$})}& $8.77$ {\scriptsize ($-1\%$)} &$1.23$& $8.89$ \\ 
 &$3840$ - $\sfrac{1}{2}$ &\multicolumn{2}{c}{\texttt{\color{gray}OOM}} &$6.46$ {\scriptsize ({\color{Maroon}$+34\%$})}& $28.75$ {\scriptsize ($-2\%$)} &$4.81$& $29.25$ \\ 
 \midrule 
\parbox[t]{20mm}{\multirow{4}{*}{DPFLow}} &$960$ - $\sfrac{1}{8}$ &$0.13$ {\scriptsize ({\color{Maroon}$+8\%$})}& $2.61$ {\scriptsize ({\color{Maroon}$+13\%$})} &$0.16$ {\scriptsize ({\color{Maroon}$+33\%$})}& $2.31$ {\scriptsize (=)} &$0.12$& $2.31$ \\ 
 &$1920$ - $\sfrac{1}{4}$ &$0.42$ {\scriptsize ({\color{Maroon}$+12\%$})}& $10.48$ {\scriptsize ({\color{Maroon}$+244\%$})} &$0.45$ {\scriptsize ({\color{Maroon}$+19\%$})}& $3.05$ {\scriptsize (=)} &$0.38$& $3.05$ \\ 
 &$3840$ - $\sfrac{1}{2}$ &\multicolumn{2}{c}{\texttt{\color{gray}OOM}} &$1.59$ {\scriptsize ({\color{Maroon}$+13\%$})}& $6.12$ {\scriptsize (=)} &$1.41$& $6.12$ \\ 
 &$7680$ - $\sfrac{1}{1}$ &\multicolumn{2}{c}{\texttt{\color{gray}OOM}} &$6.25$ {\scriptsize ({\color{Maroon}$+11\%$})}& $18.45$ {\scriptsize (=)} &$5.65$& $18.45$ \\ 
 \midrule 
\parbox[t]{20mm}{\multirow{4}{*}{SEA-RAFT}} &$960$ - $\sfrac{1}{8}$ &$0.04$ {\scriptsize ({\color{Maroon}$+2\%$})}& $2.60$ {\scriptsize ({\color{Maroon}$+17\%$})} &$0.10$ {\scriptsize ({\color{Maroon}$+175\%$})}& $2.23$ {\scriptsize (=)} &$0.04$& $2.23$ \\ 
 &$1920$ - $\sfrac{1}{4}$ &$0.17$ {\scriptsize ({\color{Maroon}$+33\%$})}& $10.07$ {\scriptsize ({\color{Maroon}$+271\%$})} &$0.21$ {\scriptsize ({\color{Maroon}$+70\%$})}& $2.71$ {\scriptsize (=)} &$0.12$& $2.71$ \\ 
 &$3840$ - $\sfrac{1}{2}$ &\multicolumn{2}{c}{\texttt{\color{gray}OOM}} &$0.77$ {\scriptsize ({\color{Maroon}$+55\%$})}& $4.66$ {\scriptsize (=)} &$0.49$& $4.66$ \\ 
 &$7680$ - $\sfrac{1}{1}$ &\multicolumn{2}{c}{\texttt{\color{gray}OOM}} &$3.13$ {\scriptsize ({\color{Maroon}$+50\%$})}& $12.46$ {\scriptsize (=)} &$2.09$& $12.46$ \\ 
 \bottomrule
\end{tabular}
\vspace*{-1mm}
\caption{
Runtime (s) and peak memory usage (GiB) of the full
optical flow method end-to-end evaluation on \textbf{\dname{Kubric-8K}} dataset
depending on the correlation computation 
variant at different scales of the inputs. We report the relative difference 
compared to our method in parentheses.
\texttt{OOM} indicates that the method requires more than $80GB$ of memory
and fails with an out-of-memory error.
}
\label{table:sup:kubric_raft_integration_results}
\end{table*}
 \begin{table*}[h!]
    \small
    \centering
    \begin{tabular}{lc@{\hspace{1em}}c@{\hspace{1em}}c@{\hspace{1em}}cccccccc}
    \toprule
    & & \multicolumn{2}{c}{\bfseries Default}
    & \multicolumn{2}{c}{\bfseries On-Demand Sampling}
    & \multicolumn{2}{c}{\bfseries Ours}
    \\
    \cmidrule(lr){3-4} \cmidrule(lr){5-6} \cmidrule(lr){7-8}
    Method & Dataset
     &  Runtime & Memory
     &  Runtime & Memory
     &  Runtime & Memory
    \\
     \midrule\parbox[t]{20mm}{\multirow{3}{*}{RAFT}}
 &\textit{KITTI 2015} &$0.08$ {\scriptsize ($-29\%$)}& $0.50$ {\scriptsize ({\color{Maroon}$+41\%$})} &$0.60$ {\scriptsize ({\color{Maroon}$+425\%$})}& $0.35$ {\scriptsize (=)} &$0.11$& $0.35$ \\ 
 &\textit{Sintel} &$0.08$ {\scriptsize ($-16\%$)}& $0.53$ {\scriptsize ({\color{Maroon}$+57\%$})} &$0.58$ {\scriptsize ({\color{Maroon}$+518\%$})}& $0.34$ {\scriptsize (=)} &$0.09$& $0.34$ \\ 
 &\textit{Spring} &$0.32$ {\scriptsize ({\color{Maroon}$+7\%$})}& $8.22$ {\scriptsize ({\color{Maroon}$+423\%$})} &$0.98$ {\scriptsize ({\color{Maroon}$+227\%$})}& $1.57$ {\scriptsize (=)} &$0.30$& $1.57$ \\ 
 \midrule 
\parbox[t]{20mm}{\multirow{3}{*}{MS-RAFT+}}
 &\textit{KITTI 2015} &\multicolumn{2}{c}{\texttt{\color{gray}OOM}} &$0.48$ {\scriptsize ({\color{Maroon}$+34\%$})}& $1.64$ {\scriptsize ($-2\%$)} &$0.36$& $1.67$ \\ 
 &\textit{Sintel} &\multicolumn{2}{c}{\texttt{\color{gray}OOM}} &$0.46$ {\scriptsize ({\color{Maroon}$+39\%$})}& $1.57$ {\scriptsize ($-2\%$)} &$0.33$& $1.60$ \\ 
 &\textit{Spring} &\multicolumn{2}{c}{\texttt{\color{gray}OOM}} &$1.68$ {\scriptsize ({\color{Maroon}$+34\%$})}& $7.01$ {\scriptsize ($-2\%$)} &$1.25$& $7.14$ \\ 
 \midrule 
\parbox[t]{20mm}{\multirow{3}{*}{DPFLow}}
 &\textit{KITTI 2015} &$0.12$ {\scriptsize ({\color{Maroon}$+11\%$})}& $0.57$ {\scriptsize ({\color{Maroon}$+63\%$})} &$0.15$ {\scriptsize ({\color{Maroon}$+39\%$})}& $0.35$ {\scriptsize (=)} &$0.11$& $0.35$ \\ 
 &\textit{Sintel} &$0.12$ {\scriptsize ({\color{Maroon}$+9\%$})}& $0.53$ {\scriptsize ({\color{Maroon}$+56\%$})} &$0.14$ {\scriptsize ({\color{Maroon}$+33\%$})}& $0.34$ {\scriptsize (=)} &$0.11$& $0.34$ \\ 
 &\textit{Spring} &$0.43$ {\scriptsize ({\color{Maroon}$+13\%$})}& $8.72$ {\scriptsize ({\color{Maroon}$+579\%$})} &$0.45$ {\scriptsize ({\color{Maroon}$+17\%$})}& $1.29$ {\scriptsize (=)} &$0.39$& $1.29$ \\ 
 \midrule 
\parbox[t]{20mm}{\multirow{3}{*}{SEA-RAFT}} 
 &\textit{KITTI 2015} &$0.04$ {\scriptsize ($-33\%$)}& $0.55$ {\scriptsize ({\color{Maroon}$+115\%$})} &$0.10$ {\scriptsize ({\color{Maroon}$+81\%$})}& $0.26$ {\scriptsize (=)} &$0.05$& $0.26$ \\ 
 &\textit{Sintel} &$0.03$ {\scriptsize (=)}& $0.58$ {\scriptsize ({\color{Maroon}$+64\%$})} &$0.09$ {\scriptsize ({\color{Maroon}$+184\%$})}& $0.34$ {\scriptsize ($-4\%$)} &$0.03$& $0.35$ \\ 
 &\textit{Spring} &$0.17$ {\scriptsize ({\color{Maroon}$+34\%$})}& $8.31$ {\scriptsize ({\color{Maroon}$+773\%$})} &$0.21$ {\scriptsize ({\color{Maroon}$+66\%$})}& $0.95$ {\scriptsize (=)} &$0.13$& $0.95$ \\ 
 \midrule 
\end{tabular}
    \caption{
    Runtime (s) and peak memory usage (GiB) of the full
    optical flow end-to-end evaluation depending on the correlation computation 
    method on three additional datasets -
    \dname{KITTI 2015}, \dname{Sintel}, and \dname{Spring}.
    }
    \label{table:sup:integration_additional_datasets}
\end{table*}
 \begin{table*}[h!]
\footnotesize
\centering
\begin{tabular}{lcccccccccc}
\toprule
& &\multicolumn{4}{c}{Metrics}&\multicolumn{4}{c}{Runtime (s) / Memory (GiB)}\\ 
\cmidrule(lr){3-6} \cmidrule(lr){7-10} 
& &1px error&EPE&LM-1px&LM-EPE&Default&On-Demand&Mixed&Ours\\ \midrule 
\parbox[t]{30mm}{\multirow{3}{*}{\begin{tabular}{@{}l@{}}MS-RAFT+~\cite{jahediMSRAFTHighResolution2024}\end{tabular}}}  & $\sfrac{1}{8}$ & $20.2$ & $1.94$ & $64.1$ & $22.57$ & \texttt{\color{gray}OOM} & $0.44\ /\ 4.4$ & $0.38\ /\ 9.2$ & $0.30\ /\ 4.4$\\ 
 & $\sfrac{1}{4}$ & $15.7$ & \underline{$\bm{1.64}$} & $41.1$ & \underline{$19.85$} & \texttt{\color{gray}OOM} & $1.40\ /\ 8.7$ & $1.35\ /\ 9.9$ & $1.04\ /\ 8.8$\\ 
 & \underline{$\sfrac{1}{2}$} & \underline{$14.5$} & $1.92$ & \underline{$34.6$} & $32.03$ & \texttt{\color{gray}OOM} & $5.47\ /\ 25.8$ & $5.45\ /\ 25.8$ & $4.11\ /\ 26.2$\\ \midrule 
\parbox[t]{30mm}{\multirow{4}{*}{\begin{tabular}{@{}l@{}}SEA-RAFT~\cite{wangSEARAFTSimpleEfficient2025}\\4LR + 0HR\\(default)\end{tabular}}}  & $\sfrac{1}{8}$ & $28.3$ & $2.30$ & $81.5$ & $21.51$ & $0.03\ /\ 3.5$ & $0.09\ /\ 3.2$ & $0.03\ /\ 3.5$ & $0.03\ /\ 3.3$\\ 
 & $\sfrac{1}{4}$ & $21.1$ & \underline{$2.01$} & $54.5$ & \underline{$19.18$} & $0.13\ /\ 8.9$ & $0.19\ /\ 3.6$ & $0.13\ /\ 8.9$ & $0.10\ /\ 3.6$\\ 
 & \underline{$\sfrac{1}{2}$} & \underline{$16.8$} & $4.94$ & \underline{$39.8$} & $39.83$ & \texttt{\color{gray}OOM} & $0.63\ /\ 5.2$ & $0.62\ /\ 5.2$ & $0.42\ /\ 5.2$\\ 
 & $\sfrac{1}{1}$ & $17.5$ & $17.65$ & $49.2$ & $107.43$ & \texttt{\color{gray}OOM} & $2.63\ /\ 11.8$ & $2.63\ /\ 11.8$ & $1.78\ /\ 11.8$\\ \midrule 
\parbox[t]{30mm}{\multirow{4}{*}{\begin{tabular}{@{}l@{}}SEA-RAFT\\3LR + 1HR\end{tabular}}}  & $\sfrac{1}{8}$ & $31.3$ & $2.67$ & $84.8$ & $27.48$ & $0.02\ /\ 3.5$ & $0.04\ /\ 3.2$ & $0.02\ /\ 3.5$ & $0.02\ /\ 3.2$\\ 
 & $\sfrac{1}{4}$ & $25.9$ & $2.25$ & $62.9$ & $20.29$ & $0.12\ /\ 8.9$ & $0.15\ /\ 3.6$ & $0.12\ /\ 8.9$ & $0.09\ /\ 3.6$\\ 
 & $\sfrac{1}{2}$ & $20.7$ & $1.93$ & $44.3$ & $18.41$ & \texttt{\color{gray}OOM} & $0.44\ /\ 5.2$ & $0.40\ /\ 5.2$ & $0.35\ /\ 5.2$\\ 
 & \underline{$\sfrac{1}{1}$} & \underline{$17.9$} & \underline{$1.83$} & \underline{$37.0$} & \underline{$17.04$} & \texttt{\color{gray}OOM} & $1.71\ /\ 11.8$ & $1.64\ /\ 11.8$ & $1.42\ /\ 11.8$\\ \midrule 
\parbox[t]{30mm}{\multirow{4}{*}{\begin{tabular}{@{}l@{}}SEA-RAFT\\6LR + 2HR\end{tabular}}}  & $\sfrac{1}{8}$ & $29.6$ & $2.42$ & $82.3$ & $22.90$ & $0.03\ /\ 3.5$ & $0.05\ /\ 3.2$ & $0.02\ /\ 3.5$ & $0.02\ /\ 3.3$\\ 
 & $\sfrac{1}{4}$ & $23.1$ & $2.03$ & $56.1$ & $17.82$ & $0.13\ /\ 8.9$ & $0.22\ /\ 3.6$ & $0.13\ /\ 8.9$ & $0.10\ /\ 3.6$\\ 
 & $\sfrac{1}{2}$ & $18.0$ & $1.84$ & $39.0$ & $17.31$ & \texttt{\color{gray}OOM} & $0.58\ /\ 5.2$ & $0.49\ /\ 5.2$ & $0.39\ /\ 5.2$\\ 
 & \underline{$\sfrac{1}{1}$} & \underline{$15.0$} & \underline{$1.82$} & \underline{$33.2$} & \underline{$17.05$} & \texttt{\color{gray}OOM} & $2.19\ /\ 11.8$ & $2.02\ /\ 11.8$ & $1.58\ /\ 11.8$\\ \midrule 
\parbox[t]{30mm}{\multirow{4}{*}{\begin{tabular}{@{}l@{}}SEA-RAFT\\4LR + 4HR\\(Cascaded)\end{tabular}}}  & $\sfrac{1}{8}$ & $28.3$ & $2.30$ & $81.5$ & $21.51$ & $0.03\ /\ 3.5$ & $0.09\ /\ 3.2$ & $0.03\ /\ 3.5$ & $0.03\ /\ 3.3$\\ 
 & $\sfrac{1}{4}$ & $21.4$ & \underline{$1.86$} & $54.1$ & \underline{$\bm{16.15}$} & $0.15\ /\ 8.9$ & $0.25\ /\ 3.6$ & $0.15\ /\ 8.9$ & $0.12\ /\ 3.6$\\ 
 & $\sfrac{1}{2}$ & $15.8$ & $1.89$ & $36.8$ & $18.55$ & \texttt{\color{gray}OOM} & $0.71\ /\ 5.2$ & $0.65\ /\ 5.2$ & $0.45\ /\ 5.2$\\ 
 & \underline{$\sfrac{1}{1}$} & \underline{$\bm{13.3}$} & $2.70$ & \underline{$\bm{31.6}$} & $21.53$ & \texttt{\color{gray}OOM} & $2.88\ /\ 11.8$ & $2.77\ /\ 11.8$ & $1.89\ /\ 11.8$\\ \bottomrule
\end{tabular}
\caption{
Analysis of alternative inference strategies on \textsc{Charge\-8K} dataset,
with larger number of iterations at low $\sfrac{1}{4}$ resolution (LR) than high resolution (HR), and mixed correlation sampler, running default implementation
up to the highest feasible scale before switching to the on-demand sampler.
}
\label{tab:supplementary:mixed_strategies}
\end{table*}
 
\FloatBarrier

\begin{listing*}[t]
\caption{Blender scene setup script.}
\label{lst:blender}
\begin{lstlisting}[language=Python,literate={"}{{\ttfamily\char34}}1]
import bpy

# Choose to render RGB or Flow pass
render_flow = False

for scene in bpy.data.scenes:

    width = 1024
    aspect_ratio = width / scene.render.resolution_x
    height = int(round(scene.render.resolution_y * aspect_ratio))

    scale = 16 if render_flow else 8
    scene.render.resolution_x = width * scale
    scene.render.resolution_y = height * scale

    scene.render.use_motion_blur = False    
    scene.render.engine = "CYCLES"
    scene.cycles.device = "CPU"

    n_samples = 1 if render_flow else 1024
    scene.cycles.samples = n_samples
    scene.cycles.adaptive_min_samples = n_samples > 1
    scene.cycles.adaptive_threshold = 0.01
    scene.cycles.use_adaptive_sampling = True
    scene.cycles.denoiser = "OPENIMAGEDENOISE"
    scene.cycles.use_denoising = (not render_flow)

    # Removing very rare but strong fireflies
    scene.cycles.sample_clamp_direct = 50.0

    # Removing lens flares and overlays
    scene.use_nodes = False

    collection = bpy.data.collections.get("flares")
    if collection:
        bpy.data.collections.remove(collection)

    for view_layer in scene.view_layers:
        view_layer.use_pass_combined = True


# Removing volumes
for mat in bpy.data.materials:
    if mat.use_nodes:
        for node in mat.node_tree.nodes:
            if node.bl_static_type == "OUTPUT_MATERIAL" and node.is_active_output:
                for link in node.inputs["Volume"].links:
                    mat.node_tree.links.remove(link)
\end{lstlisting}
\end{listing*} 

 \setlength{\tabcolsep}{4.5pt}
\begin{table*}[h!]
\footnotesize
\centering
\begin{tabular}{lccccccccc}
\toprule
&
& \multicolumn{4}{c}{Runtime (mean $\pm$ std), ms}
& \multicolumn{4}{c}{Peak Memory (mean $\pm$ std), MB}
\\ 
\cmidrule(lr){3-6} \cmidrule(lr){7-10} 
Variable & Value
& Default & On-Demand & Ours & Improvement
& Default & On-Demand & Ours & Improvement
\\ \midrule 
\parbox[t]{28mm}{\multirow{7}{*}{\begin{tabular}{@{}l@{}}Input Width\\GH200\end{tabular}}} & 32 & $27 \pm 1$ & $351 \pm 0$ & $12 \pm 1$ & $96.6\%$ & $4 \pm 0$ & $3 \pm 0$ & $3 \pm 0$ & $13.3\%$ \\ 
  & 64 & $26 \pm 1$ & $388 \pm 6$ & $12 \pm 0$ & $96.8\%$ & $27 \pm 1$ & $12 \pm 0$ & $12 \pm 0$ & $55.3\%$ \\ 
  & 128 & $25 \pm 1$ & $440 \pm 15$ & $16 \pm 1$ & $96.5\%$ & $291 \pm 0$ & $47 \pm 0$ & $46 \pm 0$ & $84.3\%$ \\ 
  & 256 & $26 \pm 1$ & $592 \pm 20$ & $26 \pm 4$ & $95.6\%$ & $4091 \pm 0$ & $184 \pm 0$ & $178 \pm 0$ & $95.6\%$ \\ 
  & 512 & $190 \pm 35$ & $1187 \pm 33$ & $94 \pm 11$ & $92.1\%$ & $71288 \pm 0$ & $737 \pm 0$ & $712 \pm 0$ & $99.0\%$ \\ 
  & 1024 & \texttt{\color{gray}OOM} & $3462 \pm 62$ & $347 \pm 32$ & $90.0\%$ & \texttt{\color{gray}OOM} & $2944 \pm 0$ & $2829 \pm 0$ & \texttt{\color{gray}-} \\ 
  & 2048 & \texttt{\color{gray}OOM} & $13522 \pm 288$ & $1334 \pm 133$ & $90.1\%$ & \texttt{\color{gray}OOM} & $11764 \pm 0$ & $11274 \pm 0$ & \texttt{\color{gray}-}\\ \midrule
\parbox[t]{28mm}{\multirow{5}{*}{\begin{tabular}{@{}l@{}}Input Width\\A100\end{tabular}}} & 32 & $25 \pm 0$ & $384 \pm 0$ & $12 \pm 0$ & $96.9\%$ & $4 \pm 0$ & $3 \pm 0$ & $3 \pm 0$ & $13.3\%$ \\ 
  & 64 & $25 \pm 0$ & $442 \pm 6$ & $20 \pm 1$ & $95.5\%$ & $27 \pm 0$ & $12 \pm 0$ & $12 \pm 0$ & $55.3\%$ \\ 
  & 128 & $25 \pm 0$ & $509 \pm 15$ & $29 \pm 3$ & $94.4\%$ & $291 \pm 0$ & $47 \pm 0$ & $46 \pm 0$ & $84.3\%$ \\ 
  & 256 & $60 \pm 0$ & $717 \pm 23$ & $58 \pm 7$ & $91.9\%$ & $4090 \pm 0$ & $184 \pm 0$ & $178 \pm 0$ & $95.6\%$ \\ 
  & 512 & \texttt{\color{gray}OOM} & $1928 \pm 73$ & $217 \pm 26$ & $88.8\%$ & \texttt{\color{gray}OOM} & $737 \pm 0$ & $712 \pm 0$ & \texttt{\color{gray}-}\\ \midrule
\parbox[t]{28mm}{\multirow{5}{*}{\begin{tabular}{@{}l@{}}Input Width\\RTX 4090\end{tabular}}} & 32 & $16 \pm 2$ & $203 \pm 0$ & $7 \pm 1$ & $96.5\%$ & $4 \pm 0$ & $3 \pm 0$ & $3 \pm 0$ & $13.3\%$ \\ 
  & 64 & $16 \pm 1$ & $242 \pm 4$ & $13 \pm 1$ & $94.8\%$ & $27 \pm 0$ & $12 \pm 0$ & $12 \pm 0$ & $55.3\%$ \\ 
  & 128 & $16 \pm 2$ & $283 \pm 11$ & $18 \pm 2$ & $93.7\%$ & $291 \pm 0$ & $47 \pm 0$ & $46 \pm 0$ & $84.3\%$ \\ 
  & 256 & $42 \pm 0$ & $394 \pm 12$ & $32 \pm 4$ & $91.9\%$ & $4090 \pm 0$ & $184 \pm 0$ & $178 \pm 0$ & $95.6\%$ \\ 
  & 512 & \texttt{\color{gray}OOM} & $1213 \pm 34$ & $121 \pm 12$ & $90.0\%$ & \texttt{\color{gray}OOM} & $737 \pm 0$ & $712 \pm 0$ & \texttt{\color{gray}-}\\ \midrule
\parbox[t]{28mm}{\multirow{5}{*}{\begin{tabular}{@{}l@{}}Input Width\\RTX 3090\end{tabular}}} & 32 & $29 \pm 0$ & $273 \pm 1$ & $13 \pm 0$ & $95.3\%$ & $4 \pm 0$ & $3 \pm 0$ & $3 \pm 0$ & $13.3\%$ \\ 
  & 64 & $29 \pm 1$ & $318 \pm 5$ & $17 \pm 1$ & $94.8\%$ & $27 \pm 0$ & $12 \pm 0$ & $12 \pm 0$ & $55.3\%$ \\ 
  & 128 & $29 \pm 2$ & $388 \pm 11$ & $30 \pm 3$ & $92.2\%$ & $291 \pm 0$ & $47 \pm 0$ & $46 \pm 0$ & $84.3\%$ \\ 
  & 256 & $79 \pm 0$ & $812 \pm 15$ & $80 \pm 6$ & $90.2\%$ & $4090 \pm 0$ & $184 \pm 0$ & $178 \pm 0$ & $95.6\%$ \\ 
  & 512 & \texttt{\color{gray}OOM} & $2615 \pm 49$ & $316 \pm 35$ & $87.9\%$ & \texttt{\color{gray}OOM} & $737 \pm 0$ & $712 \pm 0$ & \texttt{\color{gray}-}\\ \midrule
\parbox[t]{28mm}{\multirow{5}{*}{\begin{tabular}{@{}l@{}}Number of Iterations\\GH200\end{tabular}}} & 1 & $126 \pm 24$ & $37 \pm 0$ & $5 \pm 0$ & $86.3\%$ & $71288 \pm 0$ & $599 \pm 0$ & $574 \pm 0$ & $99.2\%$ \\ 
  & 8 & $140 \pm 24$ & $299 \pm 8$ & $25 \pm 2$ & $91.5\%$ & $71288 \pm 0$ & $737 \pm 0$ & $712 \pm 0$ & $99.0\%$ \\ 
  & 16 & $156 \pm 24$ & $596 \pm 16$ & $48 \pm 5$ & $91.9\%$ & $71288 \pm 0$ & $737 \pm 0$ & $712 \pm 0$ & $99.0\%$ \\ 
  & 24 & $172 \pm 24$ & $891 \pm 25$ & $71 \pm 8$ & $92.0\%$ & $71288 \pm 0$ & $737 \pm 0$ & $712 \pm 0$ & $99.0\%$ \\ 
  & 32 & $190 \pm 35$ & $1187 \pm 33$ & $94 \pm 11$ & $92.1\%$ & $71288 \pm 0$ & $737 \pm 0$ & $712 \pm 0$ & $99.0\%$\\ \midrule
\parbox[t]{28mm}{\multirow{5}{*}{\begin{tabular}{@{}l@{}}Feature Dimensionality\\GH200\end{tabular}}} & 64 & $168 \pm 25$ & $305 \pm 8$ & $63 \pm 7$ & $79.4\%$ & $71288 \pm 0$ & $599 \pm 0$ & $489 \pm 0$ & $99.3\%$ \\ 
  & 128 & $175 \pm 24$ & $598 \pm 16$ & $75 \pm 10$ & $87.4\%$ & $71288 \pm 0$ & $644 \pm 0$ & $563 \pm 0$ & $99.2\%$ \\ 
  & 256 & $190 \pm 35$ & $1187 \pm 33$ & $94 \pm 11$ & $92.1\%$ & $71288 \pm 0$ & $737 \pm 0$ & $712 \pm 0$ & $99.0\%$ \\ 
  & 512 & $224 \pm 23$ & $2362 \pm 68$ & $126 \pm 14$ & $94.7\%$ & $71288 \pm 0$ & $971 \pm 0$ & $1011 \pm 0$ & $98.6\%$ \\ 
  & 1024 & $287 \pm 21$ & $4734 \pm 145$ & $186 \pm 22$ & $96.1\%$ & $71288 \pm 0$ & $1771 \pm 0$ & $1608 \pm 0$ & $97.7\%$
\\\bottomrule
\end{tabular}
\caption{
Full correlation volume sampling isolated benchmarking results depending on one variable.
We report the runtime and peak memory usage of each implementation,
as well as memory improvement over default implementation and 
runtime improvement over the on-demand sampling method.
\texttt{OOM} indicates that the method fails with an out-of-memory error.
}
\label{table:corr_benchmark}
\end{table*}
\setlength{\tabcolsep}{6pt}
 \begin{table*}[h!]
\scriptsize
\centering
\begin{tabular}{lccccccccc}
\toprule
& &\multicolumn{4}{c}{Metrics}&\multicolumn{3}{c}{Runtime}\\ 
\cmidrule(lr){3-6} \cmidrule(lr){7-9} 
& &1px error&EPE&LM-1px&LM-EPE&Default&On-Demand&Ours\\ \midrule 
\parbox[t]{30mm}{\multirow{2}{*}{\begin{tabular}{@{}l@{}}GMFlow~\cite{xuGMFlowLearningOptical2022}\end{tabular}}}  & \underline{$\sfrac{1}{8}$} & \underline{$43.9$} & $2.95$ & $94.3$ & \underline{$24.46$} & $0.09 \pm 0.00$ & \multicolumn{2}{c}{\texttt{\color{gray}n/a}}\\ 
 & $\sfrac{1}{4}$ & $47.0$ & \underline{$2.74$} & \underline{$86.3$} & $28.61$ & $0.71 \pm 0.00$ & \multicolumn{2}{c}{\texttt{\color{gray}n/a}}\\ \midrule 
\parbox[t]{30mm}{\multirow{4}{*}{\begin{tabular}{@{}l@{}}PWC-Net~\cite{sun2018pwc}\end{tabular}}}  & $\sfrac{1}{8}$ & $42.8$ & $3.89$ & $93.6$ & $47.94$ & $0.07 \pm 0.04$ & \multicolumn{2}{c}{\texttt{\color{gray}n/a}}\\ 
 & $\sfrac{1}{4}$ & $30.3$ & $3.32$ & $78.6$ & \underline{$47.27$} & $0.10 \pm 0.05$ & \multicolumn{2}{c}{\texttt{\color{gray}n/a}}\\ 
 & \underline{$\sfrac{1}{2}$} & \underline{$24.4$} & \underline{$3.26$} & \underline{$61.4$} & $57.14$ & $0.21 \pm 0.07$ & \multicolumn{2}{c}{\texttt{\color{gray}n/a}}\\ 
 & $\sfrac{1}{1}$ & $26.0$ & $6.65$ & $76.1$ & $128.30$ & $0.64 \pm 0.10$ & \multicolumn{2}{c}{\texttt{\color{gray}n/a}}\\ \midrule 
\parbox[t]{30mm}{\multirow{4}{*}{\begin{tabular}{@{}l@{}}Flow-1D~\cite{xuHighResolutionOpticalFlow2021}\\Highres\end{tabular}}}  & $\sfrac{1}{8}$ & $36.8$ & $3.14$ & $95.2$ & $37.06$ & $0.07 \pm 0.01$ & \multicolumn{2}{c}{\texttt{\color{gray}n/a}}\\ 
 & $\sfrac{1}{4}$ & $28.2$ & $2.39$ & $83.7$ & \underline{$29.80$} & $0.23 \pm 0.05$ & \multicolumn{2}{c}{\texttt{\color{gray}n/a}}\\ 
 & \underline{$\sfrac{1}{2}$} & \underline{$23.8$} & \underline{$2.23$} & $71.3$ & $31.58$ & $0.79 \pm 0.02$ & \multicolumn{2}{c}{\texttt{\color{gray}n/a}}\\ 
 & $\sfrac{1}{1}$ & $25.6$ & $5.86$ & \underline{$69.2$} & $57.55$ & $2.82 \pm 0.01$ & \multicolumn{2}{c}{\texttt{\color{gray}n/a}}\\ \midrule 
\parbox[t]{30mm}{\multirow{4}{*}{\begin{tabular}{@{}l@{}}DIP~\cite{zhengDIPDeepInverse2022}\end{tabular}}}  & $\sfrac{1}{8}$ & $28.3$ & $2.44$ & $84.4$ & \underline{$28.92$} & $0.28 \pm 0.01$ & \multicolumn{2}{c}{\texttt{\color{gray}n/a}}\\ 
 & $\sfrac{1}{4}$ & $22.5$ & \underline{$2.23$} & $60.6$ & $33.65$ & $0.84 \pm 0.00$ & \multicolumn{2}{c}{\texttt{\color{gray}n/a}}\\ 
 & $\sfrac{1}{2}$ & $20.8$ & $2.76$ & \underline{$46.0$} & $56.97$ & $2.93 \pm 0.01$ & \multicolumn{2}{c}{\texttt{\color{gray}n/a}}\\ 
 & \underline{$\sfrac{1}{1}$} & \underline{$19.8$} & $4.00$ & $51.2$ & $85.62$ & $11.57 \pm 0.06$ & \multicolumn{2}{c}{\texttt{\color{gray}n/a}}\\ \midrule 
\parbox[t]{30mm}{\multirow{4}{*}{\begin{tabular}{@{}l@{}}FlowFormer~\cite{huangFlowFormerTransformerArchitecture2022}\end{tabular}}}  & $\sfrac{1}{8}$ & $33.2$ & $2.62$ & $89.0$ & $26.13$ & $0.79 \pm 0.00$ & \multicolumn{2}{c}{\texttt{\color{gray}n/a}}\\ 
 & $\sfrac{1}{4}$ & $25.0$ & \underline{$2.19$} & $68.0$ & \underline{$22.20$} & $1.81 \pm 0.01$ & \multicolumn{2}{c}{\texttt{\color{gray}n/a}}\\ 
 & $\sfrac{1}{2}$ & $19.2$ & $2.31$ & \underline{$52.1$} & $33.86$ & $5.06 \pm 0.03$ & \multicolumn{2}{c}{\texttt{\color{gray}n/a}}\\ 
 & \underline{$\sfrac{1}{1}$} & \underline{$16.9$} & $3.22$ & $56.4$ & $58.16$ & $16.57 \pm 0.06$ & \multicolumn{2}{c}{\texttt{\color{gray}n/a}}\\ \midrule 
\parbox[t]{30mm}{\multirow{4}{*}{\begin{tabular}{@{}l@{}}FlowFormer++~\cite{shiFlowFormerMaskedCost2023}\end{tabular}}}  & $\sfrac{1}{8}$ & $32.0$ & $2.53$ & $88.6$ & $22.44$ & $0.80 \pm 0.00$ & \multicolumn{2}{c}{\texttt{\color{gray}n/a}}\\ 
 & $\sfrac{1}{4}$ & $24.1$ & \underline{$2.00$} & $66.5$ & \underline{$20.04$} & $1.82 \pm 0.01$ & \multicolumn{2}{c}{\texttt{\color{gray}n/a}}\\ 
 & $\sfrac{1}{2}$ & $18.9$ & $2.46$ & \underline{$51.9$} & $32.26$ & $5.09 \pm 0.03$ & \multicolumn{2}{c}{\texttt{\color{gray}n/a}}\\ 
 & \underline{$\sfrac{1}{1}$} & \underline{$16.8$} & $3.41$ & $55.3$ & $55.70$ & $16.68 \pm 0.05$ & \multicolumn{2}{c}{\texttt{\color{gray}n/a}}\\ \midrule 
\parbox[t]{30mm}{\multirow{3}{*}{\begin{tabular}{@{}l@{}}SCV~\cite{jiangLearningOpticalFlow2021}\end{tabular}}}  & $\sfrac{1}{8}$ & $35.3$ & $2.60$ & $85.8$ & $29.57$ & $0.94 \pm 0.08$ & \multicolumn{2}{c}{\texttt{\color{gray}n/a}}\\ 
 & $\sfrac{1}{4}$ & $23.7$ & \underline{$2.01$} & $61.0$ & \underline{$23.19$} & $3.08 \pm 0.21$ & \multicolumn{2}{c}{\texttt{\color{gray}n/a}}\\ 
 & \underline{$\sfrac{1}{2}$} & \underline{$19.1$} & $2.83$ & \underline{$46.1$} & $46.42$ & $14.66 \pm 0.88$ & \multicolumn{2}{c}{\texttt{\color{gray}n/a}}\\ \midrule 
\parbox[t]{30mm}{\multirow{3}{*}{\begin{tabular}{@{}l@{}}HCVFlow~\cite{zhaoHybridCostVolume2024}\end{tabular}}}  & $\sfrac{1}{8}$ & $32.3$ & $2.61$ & $91.0$ & $28.68$ & $0.11 \pm 0.00$ & \multicolumn{2}{c}{\texttt{\color{gray}n/a}}\\ 
 & $\sfrac{1}{4}$ & $23.4$ & \underline{$1.99$} & $70.9$ & \underline{$25.37$} & $0.30 \pm 0.03$ & \multicolumn{2}{c}{\texttt{\color{gray}n/a}}\\ 
 & \underline{$\sfrac{1}{2}$} & \underline{$17.9$} & $2.08$ & \underline{$54.4$} & $32.92$ & $1.09 \pm 0.01$ & \multicolumn{2}{c}{\texttt{\color{gray}n/a}}\\ \midrule 
\parbox[t]{30mm}{\multirow{4}{*}{\begin{tabular}{@{}l@{}}ReCoVEr-CX~\cite{Kiefhaber_2025_ICCV}\\Mixed\end{tabular}}}  & $\sfrac{1}{8}$ & $37.7$ & $3.20$ & $97.3$ & $43.77$ & $0.04 \pm 0.00$ & \multicolumn{2}{c}{\texttt{\color{gray}n/a}}\\ 
 & $\sfrac{1}{4}$ & $26.5$ & \underline{$2.67$} & $95.1$ & \underline{$40.80$} & $0.14 \pm 0.00$ & \multicolumn{2}{c}{\texttt{\color{gray}n/a}}\\ 
 & \underline{$\sfrac{1}{2}$} & \underline{$22.6$} & $3.01$ & \underline{$94.6$} & $58.71$ & $0.56 \pm 0.00$ & \multicolumn{2}{c}{\texttt{\color{gray}n/a}}\\ 
 & $\sfrac{1}{1}$ & $23.1$ & $4.98$ & $99.6$ & $114.81$ & $2.39 \pm 0.01$ & \multicolumn{2}{c}{\texttt{\color{gray}n/a}}\\ \midrule 
\parbox[t]{30mm}{\multirow{3}{*}{\begin{tabular}{@{}l@{}}WAFT~\cite{wangWAFTWarpingAloneField2025}\end{tabular}}}  & $\sfrac{1}{8}$ & $23.3$ & $2.05$ & $83.1$ & \underline{$21.35$} & $0.16 \pm 0.03$ & \multicolumn{2}{c}{\texttt{\color{gray}n/a}}\\ 
 & $\sfrac{1}{4}$ & $16.1$ & \underline{$1.93$} & \underline{$62.0$} & $27.44$ & $0.79 \pm 0.01$ & \multicolumn{2}{c}{\texttt{\color{gray}n/a}}\\ 
 & \underline{$\sfrac{1}{2}$} & \underline{$15.3$} & $2.74$ & $62.6$ & $65.14$ & $8.47 \pm 0.07$ & \multicolumn{2}{c}{\texttt{\color{gray}n/a}}\\ \midrule 
\parbox[t]{30mm}{\multirow{3}{*}{\begin{tabular}{@{}l@{}}MS-RAFT+~\cite{jahediMSRAFTHighResolution2024}\end{tabular}}}  & $\sfrac{1}{8}$ & $20.2$ & $1.94$ & $64.1$ & $22.59$ & \texttt{\color{gray}OOM} & $0.44 \pm 0.00$ & $0.31 \pm 0.00$\\ 
 & $\sfrac{1}{4}$ & $15.7$ & \underline{$1.64$} & $41.1$ & \underline{$19.85$} & \texttt{\color{gray}OOM} & $1.41 \pm 0.01$ & $1.05 \pm 0.00$\\ 
 & \underline{$\sfrac{1}{2}$} & \underline{$14.5$} & $1.92$ & \underline{$34.6$} & $32.03$ & \texttt{\color{gray}OOM} & $5.51 \pm 0.02$ & $4.15 \pm 0.02$\\ \midrule 
\parbox[t]{30mm}{\multirow{4}{*}{\begin{tabular}{@{}l@{}}RAFT~\cite{teedRAFTRecurrentAllPairs2020}\end{tabular}}}  & $\sfrac{1}{8}$ & $32.4$ & $2.63$ & $90.3$ & $26.73$ & $0.07 \pm 0.00$ & $0.56 \pm 0.01$ & $0.09 \pm 0.00$\\ 
 & $\sfrac{1}{4}$ & $24.3$ & \underline{$2.01$} & $69.8$ & \underline{$22.16$} & $0.25 \pm 0.00$ & $0.89 \pm 0.01$ & $0.25 \pm 0.00$\\ 
 & \underline{$\sfrac{1}{2}$} & \underline{$18.9$} & $5.90$ & \underline{$49.8$} & $36.41$ & \texttt{\color{gray}OOM} & $2.62 \pm 0.03$ & $0.96 \pm 0.01$\\ 
 & $\sfrac{1}{1}$ & $25.5$ & $59.94$ & $71.3$ & $266.20$ & \texttt{\color{gray}OOM} & $10.47 \pm 0.13$ & $3.43 \pm 0.04$\\ \midrule 
\parbox[t]{30mm}{\multirow{4}{*}{\begin{tabular}{@{}l@{}}CCMR~\cite{jahediCCMRHighResolution2024}\end{tabular}}}  & $\sfrac{1}{8}$ & $25.9$ & $2.14$ & $76.1$ & \underline{$24.40$} & $0.34 \pm 0.02$ & $0.47 \pm 0.02$ & $0.33 \pm 0.01$\\ 
 & $\sfrac{1}{4}$ & $17.5$ & \underline{$1.88$} & $50.5$ & $27.71$ & \texttt{\color{gray}OOM} & $1.35 \pm 0.04$ & $1.09 \pm 0.02$\\ 
 & \underline{$\sfrac{1}{2}$} & \underline{$15.8$} & $2.16$ & \underline{$39.7$} & $42.86$ & \texttt{\color{gray}OOM} & $5.23 \pm 0.02$ & $4.39 \pm 0.01$\\ 
 & $\sfrac{1}{1}$ & $18.0$ & $4.08$ & $45.7$ & $76.96$ & \texttt{\color{gray}OOM} & $21.09 \pm 0.15$ & $17.58 \pm 0.02$\\ \midrule 
\parbox[t]{30mm}{\multirow{4}{*}{\begin{tabular}{@{}l@{}}DPFlow~\cite{morimitsuDPFlowAdaptiveOptical2025}\end{tabular}}}  & $\sfrac{1}{8}$ & $26.1$ & $2.00$ & $82.4$ & $18.88$ & $0.14 \pm 0.01$ & $0.15 \pm 0.01$ & $0.14 \pm 0.02$\\ 
 & $\sfrac{1}{4}$ & $19.6$ & \underline{$\bm{1.60}$} & $57.8$ & \underline{$\bm{13.83}$} & $0.40 \pm 0.01$ & $0.41 \pm 0.01$ & $0.35 \pm 0.01$\\ 
 & $\sfrac{1}{2}$ & $17.3$ & $1.71$ & $40.3$ & $14.29$ & \texttt{\color{gray}OOM} & $1.41 \pm 0.00$ & $1.27 \pm 0.00$\\ 
 & \underline{$\sfrac{1}{1}$} & \underline{$16.5$} & $1.92$ & \underline{$34.4$} & $18.03$ & \texttt{\color{gray}OOM} & $5.48 \pm 0.02$ & $5.00 \pm 0.02$\\ \midrule 
\parbox[t]{30mm}{\multirow{4}{*}{\begin{tabular}{@{}l@{}}SEA-RAFT~\cite{wangSEARAFTSimpleEfficient2025}\end{tabular}}}  & $\sfrac{1}{8}$ & $28.3$ & $2.30$ & $81.5$ & $21.51$ & $0.03 \pm 0.00$ & $0.09 \pm 0.00$ & $0.03 \pm 0.00$\\ 
 & $\sfrac{1}{4}$ & $21.1$ & \underline{$2.01$} & $54.5$ & \underline{$19.18$} & $0.13 \pm 0.00$ & $0.18 \pm 0.00$ & $0.11 \pm 0.00$\\ 
 & \underline{$\sfrac{1}{2}$} & \underline{$16.8$} & $4.94$ & \underline{$39.8$} & $39.83$ & \texttt{\color{gray}OOM} & $0.62 \pm 0.00$ & $0.42 \pm 0.00$\\ 
 & $\sfrac{1}{1}$ & $17.5$ & $17.65$ & $49.2$ & $107.43$ & \texttt{\color{gray}OOM} & $2.63 \pm 0.02$ & $1.78 \pm 0.01$\\ \midrule 
\parbox[t]{30mm}{\multirow{4}{*}{\begin{tabular}{@{}l@{}}SEA-RAFT\\(Cascaded)\end{tabular}}}  & $\sfrac{1}{8}$ & $28.3$ & $2.30$ & $81.5$ & $21.51$ & $0.03 \pm 0.00$ & $0.09 \pm 0.00$ & $0.03 \pm 0.00$\\ 
 & $\sfrac{1}{4}$ & $21.4$ & \underline{$1.86$} & $54.1$ & \underline{$16.15$} & $0.15 \pm 0.00$ & $0.25 \pm 0.00$ & $0.12 \pm 0.00$\\ 
 & $\sfrac{1}{2}$ & $15.8$ & $1.90$ & $36.8$ & $18.58$ & \texttt{\color{gray}OOM} & $0.72 \pm 0.00$ & $0.45 \pm 0.00$\\ 
 & \underline{$\sfrac{1}{1}$} & \underline{$\bm{13.3}$} & $2.70$ & \underline{$\bm{31.6}$} & $21.53$ & \texttt{\color{gray}OOM} & $2.89 \pm 0.02$ & $1.91 \pm 0.01$\\ \bottomrule
\end{tabular}
\vspace{-1mm}
\caption{
Full qualitative evaluation of optical flow estimation methods on the \textsc{Charge}-$8K$ dataset.
We split the results depending on the evaluation scale  and report endpoint-error (EPE), 1px outlier rate, as well as metrics for pixels with large motion.
Best result of each metric is highlighted in \textbf{bold}, best performing scale for each method is \underline{underlined}.
\texttt{OOM} indicates that the method fails with an out-of-memory error,
while \texttt{n/a} - not applicable.
We also underline the scale of the row corresponding to the results in Table 1 in the main paper.
}
\label{table:full_results}
\end{table*}
 \begin{listing*}
\begin{minted}[linenos=true, breaklines, breakafter=d, fontsize=\scriptsize]{python}
from dataclasses import dataclass
import cutlass
from cutlass import cute
from cutlass.cute.nvgpu import warp

@dataclass
class Gemm:
    """Helper class to perform tiled GEMM"""
    thr_copy: cute.TiledCopy
    rMat1: cute.Tensor
    rMat2: cute.Tensor
    rC: cute.Tensor
    tCrC: cute.Tensor
    gmem_copy: cute.TiledCopy
    smem_copy1: cute.TiledCopy
    smem_copy2: cute.TiledCopy
    tM1gM1: cute.Tensor
    tM1sM1: cute.Tensor
    tM2sM2: cute.Tensor
    mma: cute.TiledMma
    tSsMat1: cute.Tensor
    tSrMat1: cute.Tensor
    tSsMat2: cute.Tensor
    tSrMat2: cute.Tensor

    @classmethod
    def make(cls, gMat1: cute.Tensor, sMat1: cute.Tensor,
             sMat2: cute.Tensor, sCorr: cute.Tensor,
             gmem_copy: cute.TiledCopy,
             mma: cute.TiledMma):
        tidx, _, _ = cute.arch.thread_idx()
        thr_copy = gmem_copy.get_slice(tidx)
        tM1gM1 = thr_copy.partition_S(gMat1)
        tM1sM1 = thr_copy.partition_D(sMat1)
        tM2sM2 = thr_copy.partition_D(sMat2)
        thr_mma = mma.get_slice(tidx)
        rMat1 = thr_mma.make_fragment_A(thr_mma.partition_A(sMat1))
        rMat2 = thr_mma.make_fragment_B(thr_mma.partition_B(sMat2))
        rC = cute.make_fragment(thr_mma.partition_shape_C(sCorr.shape), sCorr.element_type)
        tCrC = thr_mma.partition_C(sCorr)
        smem_atom = cute.make_copy_atom(warp.LdMatrix8x8x16bOp(num_matrices=4), gMat1.element_type)
        smem_copy1 = cute.make_tiled_copy_A(smem_atom, mma)
        smem_copy2 = cute.make_tiled_copy_B(smem_atom, mma)
        s1, s2 = smem_copy1.get_slice(tidx), smem_copy2.get_slice(tidx)
        tSsMat1, tSrMat1 = s1.partition_S(sMat1), s1.retile(rMat1)
        tSsMat2, tSrMat2 = s2.partition_S(sMat2), s2.retile(rMat2)
        return cls(thr_copy, rMat1, rMat2, rC, tCrC, gmem_copy, smem_copy1, smem_copy2,
                   tM1gM1, tM1sM1, tM2sM2, mma, tSsMat1, tSrMat1, tSsMat2, tSrMat2)

    def compute_tile(self, gMat2: cute.Tensor, n_steps: int):
        tM1gM2 = self.thr_copy.partition_S(gMat2)

        self.rC.fill(0.0)

        for ko in range(n_steps):
            cute.copy(self.gmem_copy, self.tM1gM1[None, None, ko], self.tM1sM1[None, None, 0])
            cute.copy(self.gmem_copy, tM1gM2[None, None, ko], self.tM2sM2[None, None, 0])
            cute.arch.cp_async_commit_group()
            cute.arch.cp_async_wait_group(0)
            cute.arch.sync_threads()

            cute.copy(self.smem_copy1, self.tSsMat1, self.tSrMat1)
            cute.copy(self.smem_copy2, self.tSsMat2, self.tSrMat2)

            cute.gemm(self.mma, self.rC, self.rMat1, self.rMat2, self.rC)

            cute.arch.sync_threads()

        cute.autovec_copy(self.rC, self.tCrC)
        cute.arch.sync_threads()
\end{minted}
\caption{CuTe DSL GEMM helper class.}
\label{lst:kernel:a}
\end{listing*}

\begin{listing*}
\begin{minted}[linenos=true, breaklines, breakafter=d, fontsize=\scriptsize]{python}
def _floor(v: cutlass.Numeric):
    return cute.Int32(cutlass._mlir.dialects.math.floor(v.ir_value()))


def _atomicMin(p: cutlass.Pointer, v: cutlass.Int32):
    return cutlass._mlir.dialects.nvvm.atomicrmw(
        cutlass.cutlass_dsl.T.i32(),
        cutlass._mlir.dialects.nvvm.AtomicOpKind.MIN,
        p.llvm_ptr, v.ir_value(),
    )


def _map_index(i: int, h: cutlass.Constexpr, bh: cutlass.Constexpr, w: cutlass.Constexpr, bw: cutlass.Constexpr):
    bw_i = i bh_i = (i // bw) w_i = (i // (bw * bh)) h_i = i // (bw * bh * w)
    return h_i * bh + bh_i, w_i * bw + bw_i


@cute.jit
def compute_coefficients(coords, bi, by, tidx, scale, h_block, cs, r, dtype):
    ch, cw = coords.shape[2:]
    cy, cx = _map_index(by * h_block + tidx, cute.ceil_div(ch, cs), cs, cute.ceil_div(cw, cs), cs)
    is_valid_row = (cy < ch) & (cx < cw)
    x_floor, y_floor = cute.Int32(0), cute.Int32(0)
    nw, ne, sw, se = dtype(0.0), dtype(0.0), dtype(0.0), dtype(0.0)
    if is_valid_row:
        x_start = coords[bi, 0, cy, cx] * scale - r // 2
        y_start = coords[bi, 1, cy, cx] * scale - r // 2
        x_floor, y_floor = _floor(x_start), _floor(y_start)
        xc, yc = x_start - x_floor, y_start - y_floor
        nw, ne = (1 - xc) * (1 - yc), xc * (1 - yc)
        sw, se = (1 - xc) * yc, xc * yc
    return (cy, cx), is_valid_row, (x_floor, y_floor), (nw, ne, sw, se)


@cute.jit
def sample_block(sCorr, tidx, rCorr, mat2x, mat2y,
                 x_floor, y_floor, nw, ne, sw, se,
                 output, bi, cy, cx, cs, r, dtype):
    cute.autovec_copy(sCorr[tidx, None], rCorr)

    # Only iterate over parts that are in the current tile
    rx_start = max(mat2x * cs - x_floor - 1, 0)
    rx_end = min(rx_start + cs + 1, r)
    ry_start = max(mat2y * cs - y_floor - 1, 0)
    ry_end = min(ry_start + cs + 1, r)

    for ry in range(ry_start, ry_end):
        y = y_floor - mat2y * cs + ry
        for rx in range(rx_start, rx_end):
            x = x_floor - mat2x * cs + rx
            value = dtype(0.0)
            if (y >= 0) & (y < cs):
                if (x >= 0) & (x < cs):
                    value += rCorr[y * cs + x] * nw
                if (x + 1 >= 0) & (x + 1 < cs):
                    value += rCorr[y * cs + x + 1] * ne
            if (y + 1 >= 0) & (y + 1 < cs):
                if (x >= 0) & (x < cs):
                    value += rCorr[(y + 1) * cs + x] * sw
                if (x + 1 >= 0) & (x + 1 < cs):
                    value += rCorr[(y + 1) * cs + x + 1] * se
            output[bi, ry + rx * r, cy, cx] += value
\end{minted}
\caption{Kernel helper functions.}
\end{listing*}

\begin{listing*}
\begin{minted}[linenos=true, breaklines, breakafter=d, fontsize=\scriptsize]{python}
@cute.kernel
def kernel(
    coords: cute.Tensor,  # [B 2 H W]
    mat1: cute.Tensor,  # patch-major [B (h w bh bw) C]
    mat2: cute.Tensor,  # patch-major [B h w (bh bw) C]
    output: cute.Tensor,  # output [B (2r+1)^2 H H]
    sMat1_layout: cute.ComposedLayout,  # with Swizzle<2,3,3> o 0 o (8,32):(32,1)
    sMat2_layout: cute.ComposedLayout,  # with Swizzle<2,3,3> o 0 o (8,32):(32,1)
    corr_layout: cute.Layout,  # [bh, bw]
    gmem_tiled_copy_mat: cute.TiledCopy,  # with cute.nvgpu.cpasync.CopyG2SOp
    tiled_mma: cute.TiledMma,  # MmaF16BF16Op
    coord_scaler: cutlass.Numeric,  # float, scaler for coords on lower resolutions
    cell_size: cutlass.Constexpr, radius: cutlass.Constexpr, k_block: cutlass.Constexpr,
):
    (by, bi, _), (tidx, _, _) = cute.arch.block_idx(), cute.arch.thread_idx()
    h_block, w_block = corr_layout.shape[:2]
    dtype = output.element_type
    _, targetH, targetW, _, channels = mat2.shape

    smem = cutlass.utils.SmemAllocator()
    next_block = smem.allocate_tensor(cutlass.Int32, cute.make_layout((1,)))
    sMat1 = smem.allocate_tensor(mat1.element_type, sMat1_layout, byte_alignment=16)
    sMat2 = smem.allocate_tensor(mat1.element_type, sMat2_layout, byte_alignment=16)
    sCorr = cute.make_tensor(cute.recast_ptr(sMat1.iterator, dtype=dtype), corr_layout)

    gMat1 = cute.local_tile(mat1[bi, None, None], (h_block, channels), (by, 0))
    gMat1 = cute.make_tensor(gMat1.iterator.align(16), gMat1.layout)
    gemm = Gemm.make(gMat1, sMat1, sMat2, sCorr, gmem_tiled_copy_mat, tiled_mma)
    rCorr = cute.make_fragment(cute.make_layout((w_block,)), dtype)
    cute.arch.sync_threads()

    (cy, cx), is_valid_row, (x_floor, y_floor), (nw, ne, sw, se) = compute_coefficients(
        coords, bi, by, tidx, coord_scaler, h_block, cell_size, radius, dtype)

    # Find the blocks this thread requires and prepare the vote
    rp1 = radius + 1
    nr = rp1 // cell_size + 2  # Maximum number of blocks this thread can request
    blocks = cute.make_fragment(cute.make_layout((nr * nr + 1,)), cutlass.Int32)
    max_block = targetH * targetW
    blocks.fill(max_block)
    block_i = 0
    if is_valid_row:
        for i in range(nr * nr):
            rx, ry = i x = (x_floor + min(rx * cell_size, rp1 - 1)) // cell_size
            y = (y_floor + min(ry * cell_size, rp1 - 1)) // cell_size
            if (x >= 0) & (x < targetW) & (y >= 0) & (y < targetH):
                val = y * targetW + x
                if block_i == 0 or blocks[block_i - 1] < val:
                    blocks[block_i] = val
                    block_i += 1
    if tidx == 0:
        next_block[0] = max_block
    cute.arch.sync_threads()

    block_i = 0
    next_local = 0
    while next_local < max_block:
        # Vote for the next block
        if is_valid_row and blocks[block_i] < max_block:
            _atomicMin(next_block.iterator, blocks[block_i])
        cute.arch.sync_threads()
        next_local = next_block[0]

        if next_local < max_block:
            mat2y, mat2x = next_local // targetW, next_local gMat2 = cute.local_tile(mat2[bi, mat2y, mat2x, None, None], (w_block, channels), (0, 0))
            gMat2 = cute.make_tensor(gMat2.iterator.align(16), gMat2.layout)
            cute.arch.sync_threads()
            if tidx == 0:  # Resetting the vote
                next_block[0] = max_block

            gemm.compute_tile(gMat2, n_steps=(channels + k_block - 1) // k_block)
            
            if next_local == blocks[block_i]:  # If this thread requested the block, sample it and advance to the next
                sample_block(sCorr, tidx, rCorr, mat2x, mat2y, x_floor, y_floor,
                    nw, ne, sw, se, output, bi, cy, cx, cell_size, radius, dtype)
                block_i += 1
\end{minted}
\vspace*{-4mm}
\caption{Kernel implementation using CuTe DSL, reformatted for space constraints.}
\label{lst:kernel:c}
\end{listing*}

\end{document}